\pdfoutput=1
\documentclass[letterpaper]{article}
\usepackage[]{styles/acl}
\usepackage{times}
\usepackage{latexsym}
\usepackage[T1]{fontenc}
\usepackage[utf8]{inputenc}
\usepackage{microtype}
\usepackage{times,latexsym}
\PassOptionsToPackage{hyphens}{url}
\usepackage{url}
\usepackage[utf8]{inputenc}
\usepackage{color,soul}
\setul{0.45ex}{0.25ex}
\usepackage{graphicx}
\usepackage{subcaption}
\usepackage{amsmath}
\usepackage{booktabs}
\usepackage{multirow, makecell}
\usepackage{tikz}
\usepackage{graphicx}
\usepackage{array}
\usepackage{pgfplots}
\usepackage{algorithm}
\usepackage{textcomp}
\usepackage{times}
\usepackage{latexsym}
\usepackage{amsmath}
\usepackage{amsfonts}
\usepackage[noend]{algpseudocode}
\usepackage{lipsum}
\usepackage{multirow}
\usepackage{framed}
\usepackage{graphicx}
\usepackage{pifont}
\usepackage{longtable}
\usepackage{tikz}
\usepackage{booktabs}
\usepackage[all]{nowidow}
\usepackage{amssymb}
\usepackage{enumitem}
\usepackage{color,soul}
\usepackage{bm}
\usepackage{arydshln}
\usepackage{xspace}
\usepackage{siunitx}
\usepackage{booktabs}
\usepackage{booktabs}
\usepackage{xspace}
\usepackage{adjustbox}
\pgfplotsset{compat=1.14}

\newcommand{\dataset}{TCAB\xspace}

\newcommand\blankfootnote[1]{%
  \let\thefootnote\relax\footnotetext{#1}%
  \let\thefootnote\svthefootnote%
}

\title{Identifying Adversarial Attacks on Text Classifiers}

\def\authorspace{\hspace{4mm}}
\def\ucimark{$^1$}
\def\uomark{$^2$}
\def\firstauthucimark{$^{*1}$}
\def\firstauthuomark{$^{*2}$}
\author{
    Zhouhang Xie\firstauthucimark{}\authorspace{}
    Jonathan Brophy\firstauthuomark{}\authorspace{}
    Adam Noack\uomark{}\authorspace{}
    Wencong You\uomark{}\authorspace{}
    \\
    \textbf{
    Kalyani Asthana\ucimark{}\authorspace{}
    Carter Perkins\uomark{}\authorspace{}
    Sabrina Reis\uomark{}\authorspace{}
    Sameer Singh\ucimark{}\authorspace{}
    Daniel Lowd\uomark{}\authorspace{}
    }
        \\
        \ucimark{}~University of California, Irvine CA \\
        \uomark{}~University of Oregon, Eugene OR \\
        \href{mailto:zhx022@ucsd.edu}{zhx022}@ucsd.edu
        \{\href{mailto:jbrophy@cs.uoregon.edu}{jbrophy},
        \href{mailto:anoack2@cs.uoregon.edu}{anoack2}, 
        \href{mailto:wyou@cs.uoregon.edu}{wyou},
        \href{mailto:carterp@cs.uoregon.edu}{carterp},
        \href{mailto:lowd@cs.uoregon.edu}{lowd}\}@cs.uoregon.edu\\
        \href{mailto:sreis@uoregon.edu}{sreis}@uoregon.edu
        \{\href{mailto:kasthana@uci.edu}{kasthana},
        \href{mailto:sameer@uci.edu}{sameer}\}@uci.edu \\
        }

\begin{document}
\maketitle

\def\thefootnote{*}\footnotetext{Equal contribution}\def\thefootnote{\arabic{footnote}}

\begin{abstract}

The landscape of adversarial attacks against text classifiers continues to grow, with new attacks developed every year and many of them available in standard toolkits, such as TextAttack and OpenAttack.  
In response, there is a growing body of work on robust learning, which reduces vulnerability to these attacks, though sometimes at a high cost in compute time or accuracy.  In this paper, we take an alternate approach --- we attempt to understand the attacker by analyzing adversarial text to determine which methods were used to create it.  Our first contribution is an extensive dataset for attack detection and labeling: 1.5~million attack instances, generated by twelve adversarial attacks targeting three classifiers trained on six source datasets for sentiment analysis and abuse detection in English.  As our second contribution, we use this dataset to develop and benchmark a number of classifiers for attack identification --- determining if a given text has been adversarially manipulated and by which attack. As a third contribution, we demonstrate the effectiveness of three classes of features for these tasks: text properties, capturing content and presentation of text; language model properties, determining which tokens are more or less probable throughout the input; and target model properties, representing how the text classifier is influenced by the attack, including internal node activations. Overall, this represents a first step towards forensics for adversarial attacks against text classifiers.

\end{abstract}

\section{Introduction}

Text classifiers have been under attack ever since spammers started evading spam filters, nearly 20 years ago~\cite{hulten04trends}. 
In recent years, however, attacking classifiers has become much easier to carry out.
Many general-purpose attacks have been developed and are now available in standard, plug-and-play frameworks, such as TextAttack~\cite{morris2020textattack} and OpenAttack~\cite{zeng2020openattack}. 
The wide use of standard architectures and shared pretrained representations have further increased the risk of attack by decreasing the diversity of text classifiers. 

\begin{table}[tb]
\small
\setlength\tabcolsep{3pt}
\center
\begin{tabular}{lm{3.5cm}c}
\toprule
\textbf{Attack} & \textbf{Text} & \textbf{Label} \\ 
\midrule
    Original
    & the acting is amateurish
    & Negative\\
\addlinespace
    Pruthi
    & the acting is \textcolor{red}{amateirish}
    & Positive\\
\addlinespace
    DeepWordBug
    & the acting is \textcolor{red}{aateurish}
    & Positive\\
\addlinespace
    IGA
    & the acting is \textcolor{red}{enthusiastic}
    & Positive\\
\bottomrule
\end{tabular}
\caption{Attack Samples on SST-2}
\vskip -5mm
\label{tab:attack_samples_brief}
\end{table}

Our focus is on evasion attacks~\cite{barreno2006can}, in which an attacker attempts to change a classifier's prediction by making minor, semantics-preserving perturbations to the original input. 
To accomplish this, different adversarial attack algorithms employ different types of perturbations, search methods, and constraints. See Table~\ref{tab:attack_samples_brief} for some brief examples of how different attacks make different word or character substitutions to change a classifier's prediction(more examples in \S\ref{sec:attack_samples}).

A common defense strategy is to make classifiers more robust, using algorithms with heuristic or provable guarantees on their performance~\cite{madry18towards,cohen19certified}.
However, these defenses are often computationally expensive or result in reduced accuracy. 
Therefore, as a complement to making classifiers more robust, we introduce the task of \emph{attack identification} --- automatically determining the adversarial attacks (if any) used to generate a given piece of text. 
The idea behind attack identification is that many attackers will use whatever attacks are most convenient, such as public implementations of attack algorithms, instead of developing new ones or implementing ones on their own. 
Thus, we can identify specific attacks instead of detecting or preventing \emph{all} possible attacks. 
Our primary focus is on attack \emph{labeling}, 
determining which specific attack was used (or none). 
This gives us information about how the attacks are being conducted, which can be used to develop defense strategies for the overall system, such as uncovering malicious actors behind misinformation or abuse campaign on social media.

To address the problem of attack identification, we introduce the \textbf{Text Classification Attack Benchmark (TCAB)}, an extensive dataset of attacks on English text classifiers, which can be used for training and evaluating attack-identification models. 
TCAB uses six domain datasets from sentiment analysis and abuse detection. 
For each domain dataset, we train three target classifiers for the adversary to attack; we choose classifiers with transformer architectures~\cite{wolf2020transformers} as these models achieve state-of-the-art results \emph{and} previous work has shown that adversarial examples curated against transformers have the highest transferability to other architectures such as CNNs and LSTMs, while the reverse is not true~\cite{Li2021searching}.
We then run twelve attacks from the TextAttack and OpenAttack toolkits against each classifier for all datasets, this amounts to a total of 216 domain dataset/target model/attack combinations. The final TCAB dataset consists of: (1) all attacks that successfully flipped the label of the target classifier, a total of 1,539,881 adversarial instances, and (2) the unperturbed ``clean'' instances from the original domain test sets. 
The data and code used to generate TCAB will be released publicly, so it can be expanded and updated with additional datasets and new attacks as they are developed.

To characterize the text properties that are useful in identifying attacks, we investigate three classes of features. 
\emph{Text properties (T)} capture the content and presentation of the text, such as the 
number of non-ASCII characters and average word length. 
This can differentiate attacks that perform different transformations on the input, from replacing words with synonyms to inserting characters or punctuation. \emph{Language model properties (L)} indicate how natural each token is in 
the context of the input, helping to identify attacks that, for example, generate very improbable tokens. 
\emph{Target classifier properties (C)} represent how a text classifier responds to the input, including its internal node activations and gradients. We refer to these three sets of features collectively as \emph{TLC features}. 
We combine these with a standard BERT~\cite{devlin2019bert} representation of the input.

We evaluate linear and tree-ensemble classifiers on TCAB using this rich feature set. We find that attack detection (determining if any attack is present) can be done with 84--97\% accuracy when the target model used to generate the attack matches the target model properties, and  83--91\% accuracy when they differ. 
We also find that our models generalize to detecting attacks that were unseen at training time, which is promising for detecting new attacks in the wild. For labeling, accuracies range from  45--71\% accuracy\footnote{Test sets are balanced, so baseline accuracy is 50\% for binary detection and 8.33\% on the multiclass labeling task.}. 
Together, the TCAB dataset, our TLC features, and the evaluation represent a substantial first step towards automated identification of adversarial attacks against text classifiers.

\section{Attack Identification}
\label{sec:problem}

Existing work on defending against adversarial textual attacks mainly focuses on building robust models via adversarial training~\cite{miyato2017adversarial}, and much less attention has been put on detecting the presence of an attack~\cite{mozes-etal-2021-frequency}~(for a detailed list of existing defenses, see~\S\ref{sec:related_work}). In computer vision, recent work has shown that attacks on image classifiers can be detected and the attack method responsible can be identified~\cite{moayeri2021sample}; they argue that knowing which specific attack was used allows for more specific defenses. 
With an increasing number of published textual attack methods~\cite{gao2018deepwordbug,ramakrishnan2020bae,zang2020pso} and attack frameworks~\cite{morris2020textattack,zeng2020openattack}, being able to not only detect an attack but also \emph{label} the attack method used to perturb the input may better help NLP practitioners understand the attack.

Strictly speaking, labeling is not necessary for defending classifiers --- if we determine with some confidence that an attack is present, we can simply label a piece of text (or its source) as malicious. The benefit of labeling is to learn more about the attacker. For example, if we find that the same attack is being applied by many different accounts on a social network, then we may reasonably suspect that all are part of a coordinated bot attack. Alternatively, if we find that an attack is novel and does not match any we have seen before, then we might suspect the attacker to be more sophisticated and have more resources than an attacker using a standard implementation. Thus, the information gained from attack labeling could be used when developing defense strategies for the overall system.

\paragraph{Problem Setup}
In this work, we focus on text classifiers and attacks on them. 
Given an input sequence $\mathbf{x} = (x_1, x_2, ..., x_N) \in \mathcal{X}$~(the instance space), a text classifier $f$ maps $\mathbf{x}$ to a label $y \in \mathcal{Y}$, the set of output labels.
For sentiment analysis, $\mathcal{Y}$ may be positive or negative sentiment; or for toxic comment detection, $\mathcal{Y}$ may be toxic or non-toxic.

A text-classification adversary aims to generate an adversarial example $\mathbf{x'}$ such that $f(\mathbf{x'}) \neq f(\mathbf{x})$. Ideally, the changes made on $\mathbf{x}$ to obtain $\mathbf{x'}$ are minimal such that a human would label them the same way. Perturbations may occur on the character-, token-, phrase-, or sentence-level, or a combination of levels; perturbations may also be structured such that certain input properties are preserved, such as the semantics, perplexity, fluency, or grammar.

Given a (possibly) perturbed input sequence $\mathbf{x^*} \in \mathcal{X}$, we aim to develop a detector $f^D: \mathcal{X} \rightarrow \{-1, +1\}$ that detects the presence of any perturbations on $\mathbf{x^*}$; we call this task \emph{attack detection}. In addition, we aim to identify the method used to perturb the input. Given $\mathbf{x^*} \in M(\mathbf{x})$, in which $M(\mathbf{x})$ is a function that perturbs $\mathbf{x}$ using any one attack method from a set of attacks $S$ (including a ``clean'' attack in which the input is not perturbed), we develop an attack labeler $f^L: \mathcal{X} \rightarrow S$; we introduce this more difficult task as \emph{attack labeling}.

In pursuit of these aims, we develop and curate a large collection of adversarial attacks on a number of classifiers trained on various domain datasets. In the following sections, we describe our process for generating this benchmark, and then detail the features we use to build $f^D$ and $f^L$. Finally, we evaluate the effectiveness of our models in detecting and labeling different attack methods.

\section{Creating an Identification Benchmark}

We now present the Text Classification Attack Benchmark (TCAB), a dataset for developing and evaluating methods for identifying adversarial attacks against text classifiers.

\subsection{Tasks and Domain Datasets}

For sentiment analysis, we attack models trained on three domains: 
(1) \textbf{Climate Change}\footnote{\url{https://www.kaggle.com/edqian/twitter-climate-change-sentiment-dataset}}, 62,356 tweets on climate change; (2) \textbf{IMDB}~\cite{maas-EtAl:2011:ACL-HLT2011}, 50,000 movie reviews,
and (3) \textbf{SST-2}~\cite{socher2013recursive}, 68,221 movie reviews.
For abuse detection, we attack models trained on three toxic-comment datasets: (1)~\textbf{Wikipedia} (Talk Pages)~\cite{wulczyn2017ex, dixon2018measuring}, which contains 159,686 comments from Wikipedia administration webpages, (2) \textbf{Hatebase}~\cite{davidson2017automated}, which contains 24,783 comments, and (3) \textbf{Civil Comments}\footnote{https://www.kaggle.com/c/jigsaw-unintended-bias-in-toxicity-classification}, which contains 1,804,874 comments from independent news sites. 
All datasets are binary (positive vs.\ negative or abusive vs.\ non-abusive) except for Climate Change, which includes neutral sentiment.

To create TCAB, we perturb examples from the test sets of these six domain datasets. SST-2, Wikipedia (Talk Pages), and IMDB
have predefined train/test splits.
For the other 
three datasets, we use an 80/10/10 split for training, validation, and testing. For each model/domain dataset combination, we only attack test set examples in which the model's prediction is correct. For the abuse datasets, we further constrain our focus to examples in the test set that are both predicted correctly \emph{and} toxic; perturbing non-toxic text to be classified as toxic is a less likely adversarial task.

\subsection{Target Models}

\begin{table}[tb]
\small
\setlength\tabcolsep{3pt}
\center
\begin{tabular}{l ll ll ll}
\toprule
 & \multicolumn{2}{c}{\textbf{BERT}}
 & \multicolumn{2}{c}{\textbf{RoBERTa}}
 & \multicolumn{2}{c}{\textbf{XLNet}}
\\
 \cmidrule(lr){2-3}
 \cmidrule(lr){4-5}
 \cmidrule(lr){6-7}
 \textbf{Dataset} & Acc. & AUC & Acc. & AUC & Acc. & AUC \\
\midrule
Climate Change*    & 79.8 & 0.899 & \textbf{81.2} & \textbf{0.917} & 80.1 & 0.910 \\
IMDB               & 87.0 & 0.949 & \textbf{90.7} & \textbf{0.968} & 90.1 & 0.965 \\
SST-2                & 91.8 & 0.972 & \textbf{92.7} & \textbf{0.978} & 92.3 & 0.974 \\
Wikipedia          & 96.5 & 0.982 & \textbf{96.6} & \textbf{0.985} & 96.4 & 0.983 \\
Hatebase           & \textbf{95.8} & 0.983 & \textbf{95.8} & \textbf{0.987} & 93.9 & 0.979 \\
Civil Comments     & \textbf{95.2} & \textbf{0.968} & 95.1 & 0.967 & 95.0 & 0.965 \\
\bottomrule
\end{tabular}
\caption{Predictive performance of the target models on test set for each domain dataset; *: multiclass-macro-averaged AUC; the rest are binary-classification tasks.}
\label{tab:victim_model_performance}
\end{table}

We finetune BERT, RoBERTa, and XLNet models — all from HuggingFace's transformers library~\cite{wolf2020transformers} — on the six domain datasets. Table~\ref{tab:victim_model_performance} shows the performance of these models on the test set of each domain dataset. On most datasets, RoBERTa slightly outperforms the other two models both in accuracy and AUROC. We will make all target models publicly available.

\begin{figure*}[t]
    \centering
    \includegraphics[width=\textwidth]{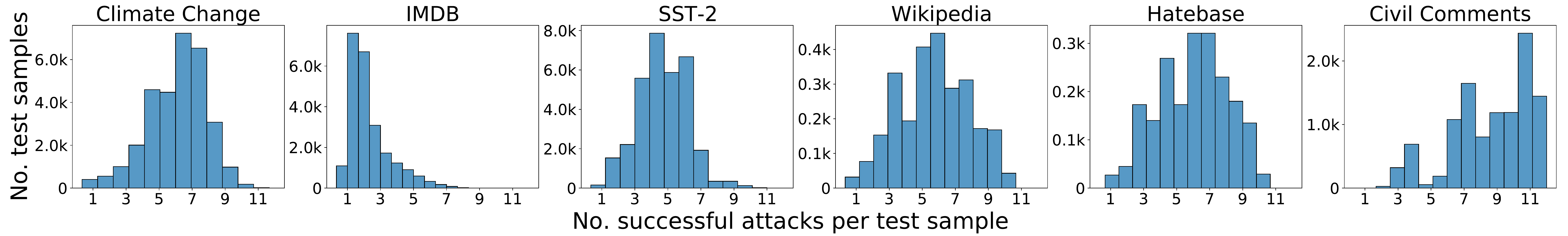}
    \caption{Histogram for number of successful attacks (out of 12), averaged across all three target models.}
    \label{fig:success_dists}
\end{figure*}

\subsection{Attack Methods}
We attempt to attack all target models using all of the attack methods implemented in two publicly available and easy-to-use toolchains: TextAttack~\cite{morris2020textattack} and OpenAttack~\cite{zeng2020openattack}. 
Of the 16 attack methods offered by TextAttack, we find eight consistently fool the target models without crashing:
(1) BAE~\cite{ramakrishnan2020bae},
DeepWordBug~\cite{gao2018deepwordbug}, FasterGenetic (a modified version of the attack proposed in~\citet{jia2019certified}), IGA~\cite{wang2019igawang}, Pruthi~\cite{pruthi2019combating}, PSO~\cite{zang2020pso}, TextBugger~\cite{li2018textbugger}, and TextFooler~\cite{jin2019textfooler}. BAE, FasterGenetic, IGA, PSO, and TextFooler perturb text at the word level, whereas DeepWordBug, Pruthi, and TextBugger all perturb text at the character level. 
Of the 13 attacks implemented by OpenAttack, we find only four consistently fool the target models without crashing: Genetic~\cite{alzantot2018genetic}, HotFlip~\cite{ebrahimi2018hotflip}, TextBugger~\cite{li2018textbugger}, and VIPER~\cite{eger2019viper}. Genetic perturbs text at the word level, HotFlip and TextBugger perturb text at the word level and at the character level, and VIPER perturbs text at the character level. See Table \ref{tab:attack_methods} for a taxonomy of all 12 attack methods used to create TCAB.

\subsection{\dataset Statistics}

\dataset consists of 1,539,881 successful attacks,
and Table \ref{tab:success-rates} shows a breakdown of attack success rates and number of successful attacks for each method.

\begin{figure}[tb]
    \setlength{\belowcaptionskip}{-5pt}
    \centering
    \includegraphics[width=.46\textwidth]{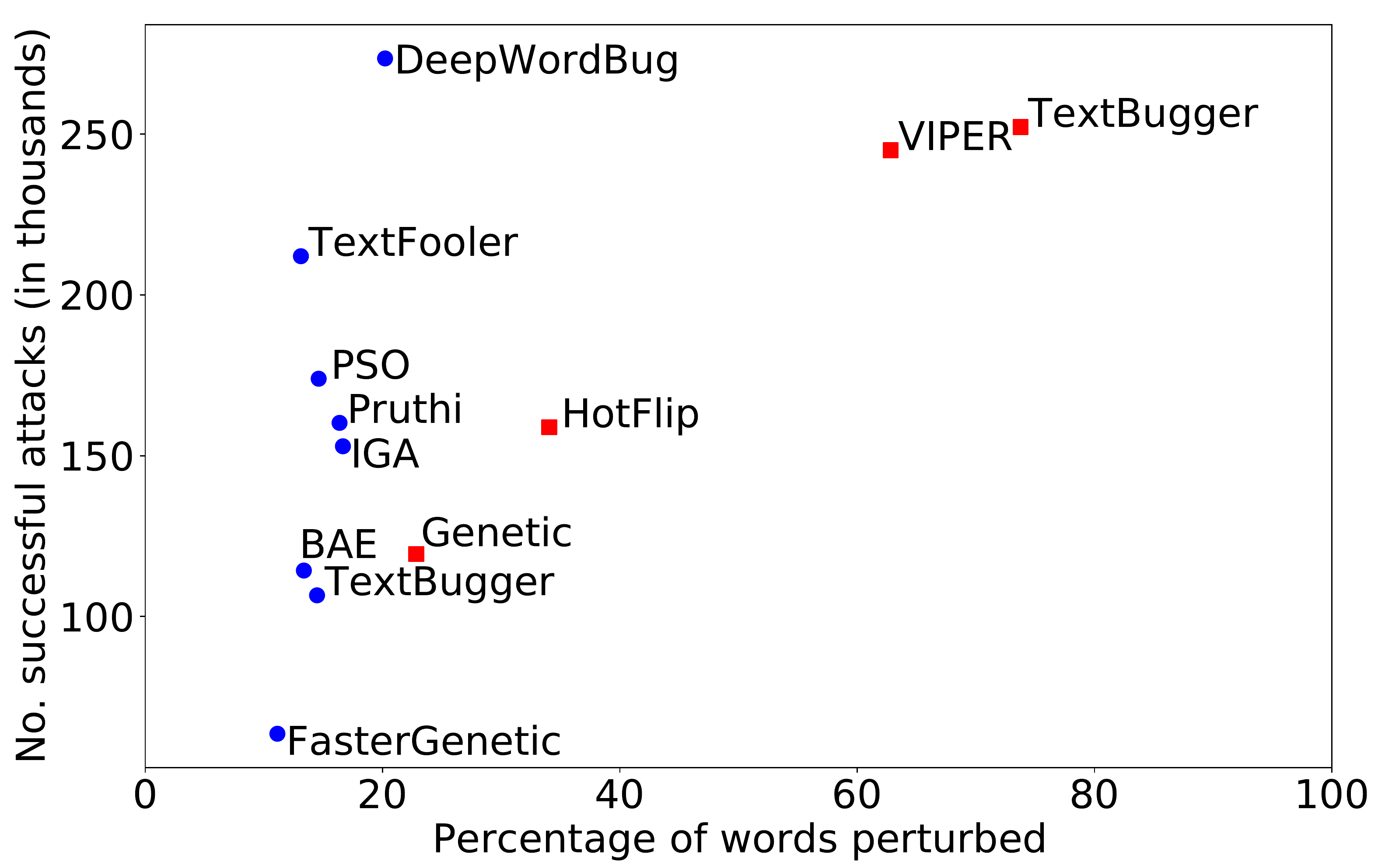}
    \caption{Average percentage of words perturbed per successful attack vs. the number of successful attacks across all domains, with TextAttack as blue circles and OpenAttack as red squares.}
    \label{fig:pert_perc_vs_num_success}
    
\end{figure}

Interestingly, the degree of input perturbation and attack-success frequency are only somewhat correlated~(Figure~\ref{fig:pert_perc_vs_num_success}). For example, DeepWordBug perturbs just 20\% of the input words on average, but generates more successful attacks than any other method. The four attack methods from OpenAttack perturb more words in the input than any of the TextAttack methods. 

For Civil Comments, many instances were very easy to manipulate successfully~(Figure \ref{fig:success_dists}: far right), and it was not uncommon for all 12 attackers to successfully perturb the same instance. In contrast, it was quite rare for more than three of the attackers to be successful on any IMDB instance. 
Of the three target-model architectures, XLNet was the most robust — it was successfully attacked 57\% of the time (this percentage is an average across all attack attempts made against all XLNet models). BERT and RoBERTa were similar in robustness, both being fooled 60\% of the time.

\begin{table*}[t]
\setlength{\tabcolsep}{5pt}
\small
\center
\begin{tabular}{lcccccc}
\toprule
\textbf{Attack Method} & \textbf{Clim. Cha.} & \textbf{IMDB} & \textbf{SST-2} & \textbf{Wikipedia} & \textbf{Hatebase} & \textbf{Civ. Com.} \\
 \midrule
\textbf{BAE}~\cite{ramakrishnan2020bae} &   52 (21.4k)  &  36 (\hphantom{0}5.1k)  &68 (\hphantom{0}4.0k)  &61 (4.5k)  &61 (3.6k)  &71 (21.6k)   \\
DeepWordBug~(\textbf{DWB})~\cite{gao2018deepwordbug} &  86 (80.9k)  &  74 (16.6k)  &  79 (74.1k)  &  79 (5.8k)  &76 (4.6k)  &99 (30.0k)   \\
FasterGenetic~(\textbf{FG})~\cite{jia2019certified} &   38 (14.1k)  &  11 (\hphantom{0}2.2k)  &30 (\hphantom{0}1.8k)  &32 (2.4k)  &33 (2.0k)  &66 (19.9k)   \\
Genetic*~(\textbf{Gn.*})~\cite{alzantot2018genetic} &   67 (24.2k)  &  46 (11.7k)  &  34 (29.5k)  &  45 (3.4k)  &13 (0.8k)  &80 (24.3k)  \\
HotFlip*~(\textbf{HF*})~\cite{ebrahimi2018hotflip} &    52   (49.5k)  &  36 (12.3k)  &  42 (39.5k)  &  37 (2.8k)  &35 (2.1k)  &75 (22.6k)  \\
\textbf{IGA}~\cite{wang2019igawang}  &  52 (49.1k)  &  \hphantom{0}0 (\hphantom{0.0k}0) & 59 (54.3k)  &  \hphantom{0}0 (\hphantom{0.k}0) & 54 (3.7k)  & 64 (19.3k) \\
Pruthi~(\textbf{Pr.})~\cite{pruthi2019combating} &  43 (40.7k)  &  19 (\hphantom{0}5.2k)  &59 (55.9k)  &  35 (2.6k)  &40 (2.4k)  &67 (20.3k)  \\
\textbf{PSO}~\cite{zang2020pso} &   59 (55.7k)  &  27 (\hphantom{0}9.3k)  &72 (66.7k)  &  31 (2.3k)  &35 (2.1k)  &62 (18.7k)  \\
TextBugger*~(\textbf{TB*}) \cite{li2018textbugger} &    81 (75.9k)  &  79 (33.1k)  &  65 (61.4k)  &  74 (5.5k)  &54 (3.2k)  &97 (29.3k)  \\  
TextBugger~(\textbf{TB})~\cite{li2018textbugger} &  74 (\hphantom{0}6.3k)  &57 (\hphantom{0}8.0k)  &68 (\hphantom{0}4.0k)  &65 (4.8k)  &56 (3.4k)  &95 (28.8k)  \\
TextFooler~(\textbf{TF})~\cite{jin2019textfooler} & 92 (86.5k)  &  51 (\hphantom{0}9.9k)  &94 (\hphantom{0}5.5k)  &82 (6.0k)  &83 (5.0k)  &98 (29.5k)  \\
VIPER*~(\textbf{VIP*})~\cite{eger2019viper} &   62 (58.7k)  &  88 (38.8k)  &  63 (59.6k)  &  66 (4.9k)  &67 (4.0k)  &75 (22.8k)  \\
\bottomrule
\end{tabular}
\caption{Percentage (and number) of successful attacks across all three target models. Attack methods with an ``*'' are from the OpenAttack toolchain, those without are from the TextAttack toolchain.}
\label{tab:success-rates}
\end{table*}

\section{Feature Sets for Attack Identification}

\begin{figure}[t]
    \centering
    \includegraphics[width=.47\textwidth]{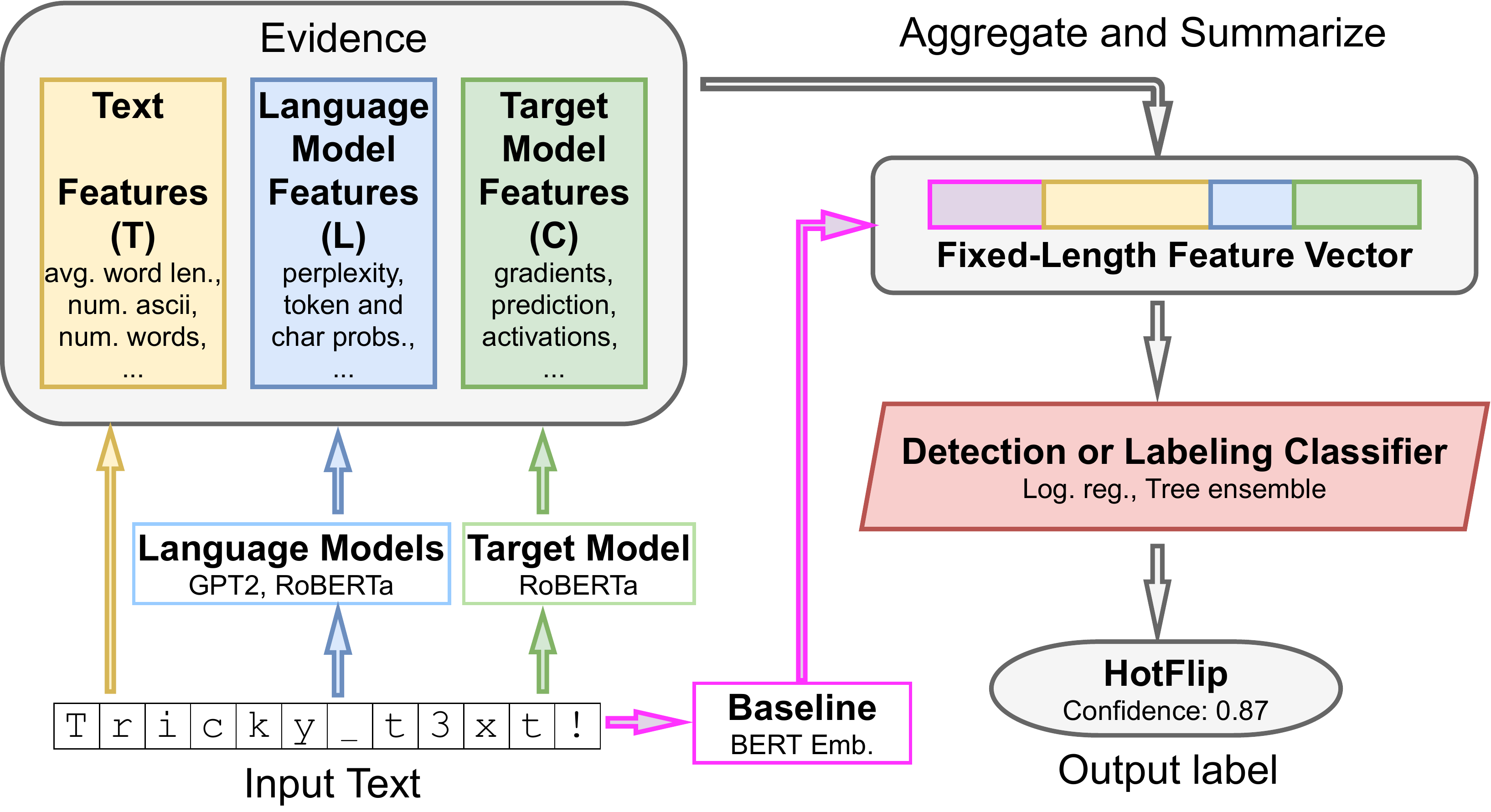}
    \caption{Pipeline for attack detection and labeling.}
    \label{fig:architecture}
\end{figure}

At a high level, our pipeline (Figure~\ref{fig:architecture}) extracts 
three types of
features from the input (T, L, C), aggregates them into a fixed-length vector, and uses a trained classifier to 
identify determine the most probable attack method %
(\emph{labeling}) or the probability that the instance is adversarial (\emph{detection}).

\paragraph{Text Properties, T}

We use BERT~\cite{devlin2019bert} to generate contextualized embeddings of the input as a baseline set of features (c.f.~\cite{zhou2019learning}).
Second, we extract features from the input text such as length, non-ascii character count, token casing/shape, punctuation marks, and other surface-level characteristics that may have been modified by the attacks. 

\paragraph{Language Model Properties, L}

We compute the probability and rank of each token using RoBERTa, and the perplexity of the input sequence using GPT-2~\cite{radford2019language}. 
These features identify structures in the language of the input text, such as ungrammatical, awkward, or generic phrasing.

\paragraph{Target Model Properties, C}

We use the target model's output posteriors, node activations, gradients, and saliency~(gradients w.r.t. input tokens) to capture any changes in the target model due to deceptive input.
This measures the effect of deceptive text on the target classifier.

\paragraph{Aggregation} 
For token-level properties, such as language model probability and rank, we compute mean, variance, and quantiles, both across the entire input sequence and within different input regions~(first 25\%, middle 50\%, last 75\%). This gives us a fixed-length feature vector. For a complete and detailed list of all properties, see~\S\ref{sec_appendix:samplewise_properties}.

\section{Experimental Evaluation}

We conduct extensive evaluations on attack detection and labeling to determine the effectiveness of our proposed TLC features and the relative difficulty of identifying attacks in the TCAB dataset.

\subsection{Setup}

We apply weight balancing and oversampling so that each class in the TCAB dataset is equally represented.
Since the TextBugger attack was implemented by both TextAttack and OpenAttack, we merge the instances to consider them a single attack, resulting in 11 attacks.
As baselines, 
we train a standard logistic regression classifier
and a gradient-boosting tree ensemble~\cite[LightGBM]{ke2017lightgbm}
applied to BERT features. 
We also train these classifiers
with different combinations of our features, that is:
\begin{itemize}[nosep,leftmargin=12pt]
     \item L/T: baseline \textbf{L}inear or \textbf{T}ree ensemble model with BERT representation.
     \item L/T-T: includes hand-crafted \textbf{T}ext properties.
     \item L/T-TL: includes \textbf{L}anguage model properties.
     \item L/T-TLC: includes target \textbf{C}lassifier properties.
\end{itemize}

\begin{figure*}[t]
\begin{subfigure}{\textwidth}
    \centering
    \caption{Attack Detection (2 classes)}
    \hspace*{-.28cm}  
    \includegraphics[width=1.01\textwidth,clip,trim=0 0 0 0]{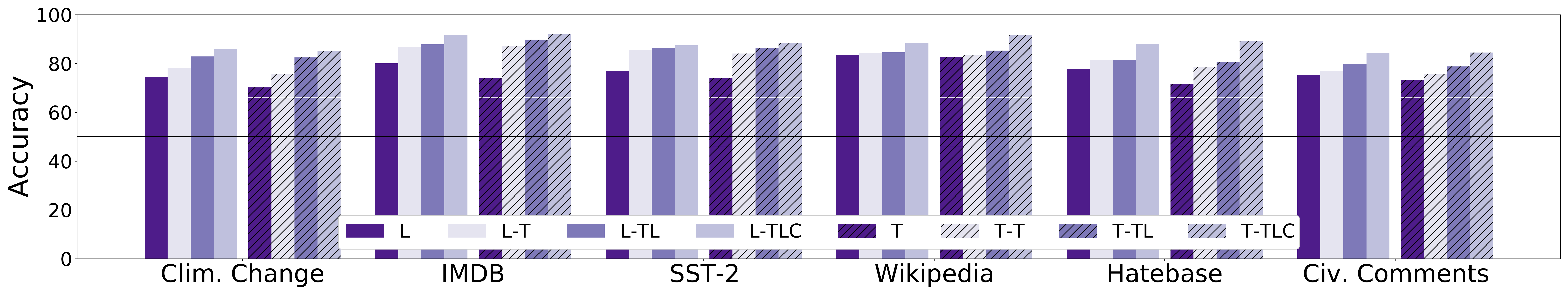}
    \label{fig:clean_vs_all}
\end{subfigure}
\begin{subfigure}{\textwidth}
    \centering
    \caption{Attack Labeling (12 classes)} 
    \includegraphics[width=\textwidth,clip,trim=0 0 0 0]{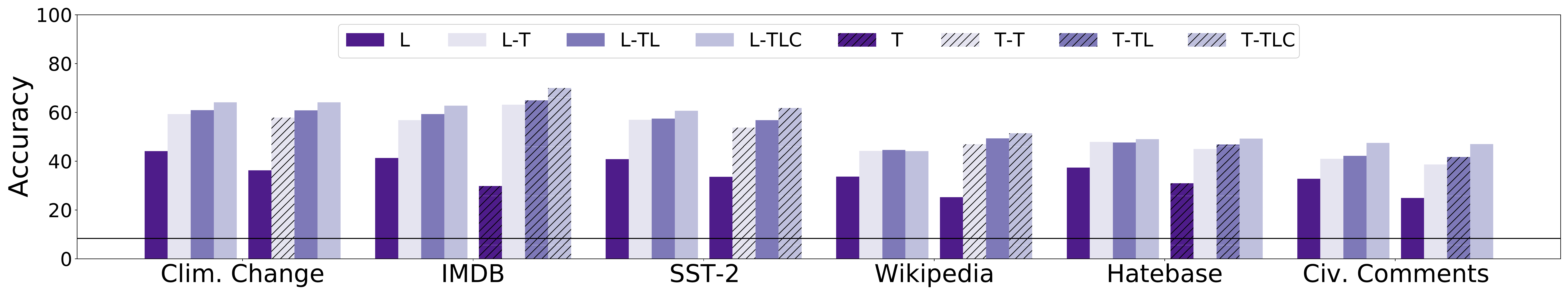}
    \label{fig:multiclass_with_clean}
\end{subfigure}
    \vskip -5mm
    \caption{\textbf{Attack detection and labeling} results averaged across the three target classifiers: BERT, RoBERTa, and XLNet (see Tables \ref{tab:clean_vs_all} and \ref{tab:mulitclass_with_clean} in the Appendix for unaggregated accuracy numbers). Horizontal lines represent random baseline performance.}
    \label{fig:detection_and_labeling_bars}
        \vskip -5mm
\end{figure*}

\begin{figure}[t]
    \centering
        \includegraphics[width=1.0\columnwidth]{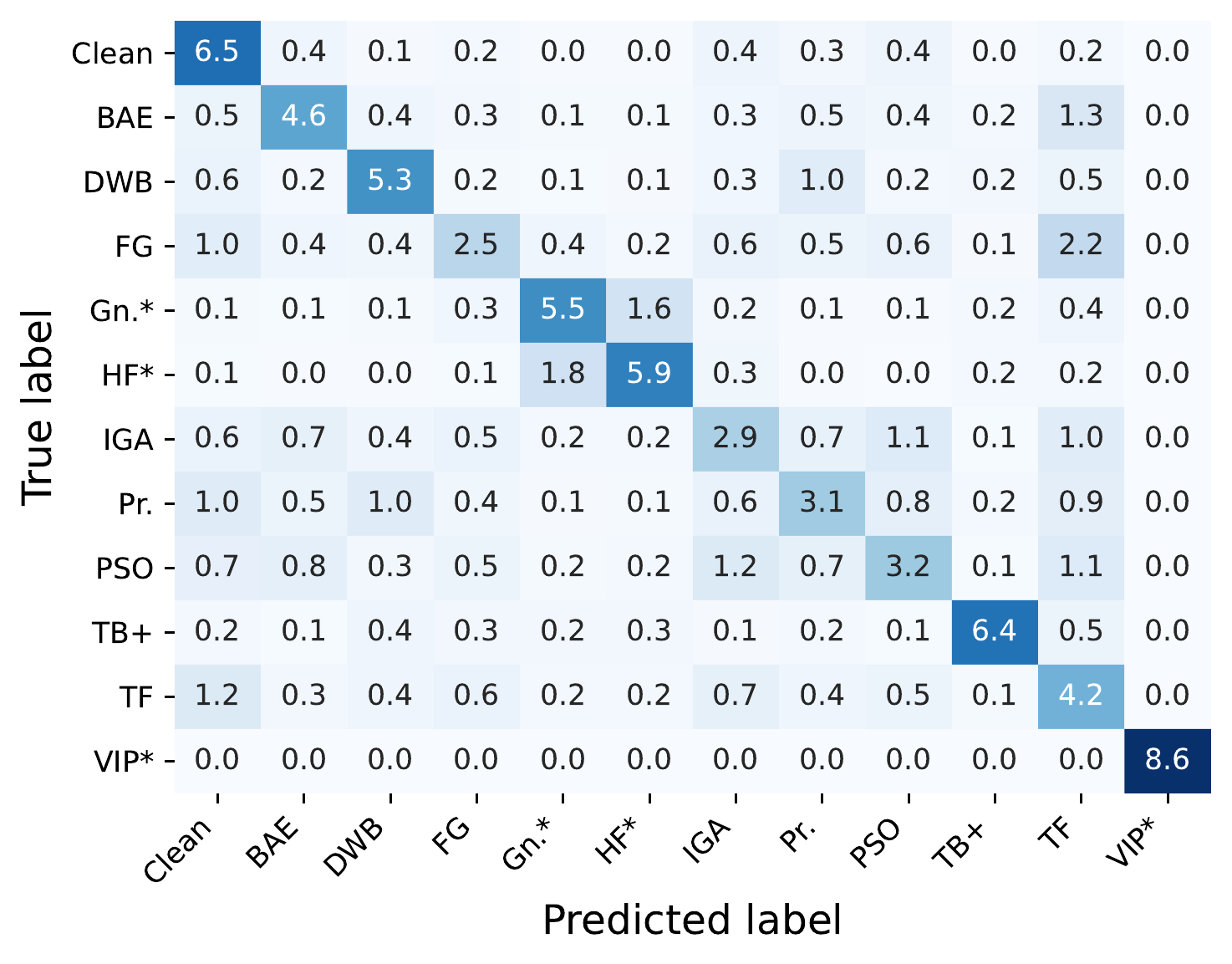}
    \caption{Confusion matrix results using T-TLC to label attacks; each entry represents the percentage of predictions made, averaged over all target models and datasets; confusion matrices for each dataset are in ~\S\ref{appendix_sec:confusion_matrix}. 
    }
        \vskip -5mm
    \label{fig:confusion_matrix}
\end{figure}

\subsection{Detection and Labeling Methods}
\label{detection}
Considering the scenario where the attack methods, domain tasks, and target models are all known ahead of time, we test the ability of our methods to identify the presence of any attack in two categories: binary classification of separating perturbed texts versus the clean ones; and multiclass labeling in which each label corresponds to either a different attack or the original clean text. 

When separating all attacks from clean data~(Figure~\ref{fig:clean_vs_all}, complete results in the Appendix, Table~\ref{tab:clean_vs_all}), we see strong results across all target classifiers.
T shows a small improvement over L in most cases; additionally, L and T with TLC features outperform other feature ablations in nearly all cases, and outperform the BERT baseline (L and T) by 8\% and 13\% on average, respectively.
When labeling clean data and every attack, we see substantial improvement over the baseline~(Figure~\ref{fig:multiclass_with_clean}, complete results are in the Appendix, Table~\ref{tab:mulitclass_with_clean}). T continues to show a small improvement over L due to its greater flexibility. L and T with language model and target model features outperform the BERT baseline (L and T) by 20\% and 23\% on average, respectively.

Figure~\ref{fig:confusion_matrix} shows the confusion matrix of predictions for the attack labeling task.
We observe word-level attacks such as HF*, PSO, and BAE are easier to distinguish than most character-level attacks, due to the effectiveness of the language-model features.
This finding aligns with our results in Figure~\ref{fig:multiclass_with_clean}
where text features yield the least accurate predictions among all ablations. The one exception is 
VIP*, a character-level attack that uses a unique visual character embedding method, which is easy to detect and label.
We also observe the model tends to confuse GN* and HF* attacks.
On average, FG, Pr., and PSO tend to be the most difficult attacks to detect across all datasets.
Confusion matrices for each individual dataset are in the Appendix,~\S\ref{appendix_sec:confusion_matrix}.

\subsection{Generalizing to New Target Model}

We test the ability of our methods to detect and label attacks when the target model is \emph{not} known ahead of time by training our detection/labeling models on attacks aimed at two out of the three possible target classifiers~(BERT, RoBERTA, XLNet) and evaluating them on attacks aimed at the remaining held-out target model. We compare the performance of L and T models with TLC features. Table~\ref{tab:heldout_target_model} shows our methods generalize well to unseen target models for both attack detection~(85\% and 87\% acc.\ for L and T, on average) and attack labeling~(54\% and 57\% acc.\ for L and T). %

\begin{table*}[t]
\small
\setlength\tabcolsep{3pt}
\center
\begin{tabular}{l cc cc cc c cc cc cc}
\toprule
\multirow{2}{*}{}
  & \multicolumn{6}{c}{\textbf{Attack Detection} (2 classes)}
  &
  & \multicolumn{6}{c}{\textbf{Attack Labeling} (12 classes)} \\
\cmidrule(lr){2-7}
\cmidrule(lr){9-14}
  \textbf{Held-Out Target}: & \multicolumn{2}{c}{\textbf{BERT}} & \multicolumn{2}{c}{\textbf{RoBERTA}} & \multicolumn{2}{c}{\textbf{XLNet}} &
  &
  \multicolumn{2}{c}{\textbf{BERT}} & \multicolumn{2}{c}{\textbf{RoBERTA}} & \multicolumn{2}{c}{\textbf{XLNet}} \\
\cmidrule(lr){2-3}
\cmidrule(lr){4-5}
\cmidrule(lr){6-7}
\cmidrule(lr){9-10}
\cmidrule(lr){11-12}
\cmidrule(lr){13-14}
 \textbf{Dataset} & L\scriptsize-TLC &
 T\scriptsize-TLC &   L\scriptsize-TLC &
 T\scriptsize-TLC &   L\scriptsize-TLC &
 T\scriptsize-TLC & & L\scriptsize-TLC &
 T\scriptsize-TLC &   L\scriptsize-TLC &
 T\scriptsize-TLC &   L\scriptsize-TLC &
 T\scriptsize-TLC \\
\midrule
Climate Change
               & \textbf{83.5} & 83.1
               & \textbf{86.4} & 86.0
               & \textbf{82.8} & 81.7
               &
               & \textbf{56.1} & 55.5
               & 63.5          & \textbf{64.6}
               & 59.2          & \textbf{59.5}
\\
IMDB
      & 88.1 & \textbf{89.4}
      & 80.1 & \textbf{82.3} 
      & \textbf{87.6} & 85.2
      & \
      & 59.6 & \textbf{63.8}
      & 60.5 & \textbf{63.1}
      & 52.3 & \textbf{52.9}
\\
SST-2
      & 88.5 & \textbf{88.9}
      & 88.5 & \textbf{88.7}
      & 85.6 & \textbf{87.4}
      &
      & 60.4          & \textbf{60.9}
      & \textbf{61.1} & 60.8
      & 60.3          & \textbf{61.7}
\\
Wikipedia
      & 89.0 & \textbf{92.4}
      & 87.6 & \textbf{89.4}
      & 87.9 & \textbf{90.8}
      &
      & 49.2 & \textbf{54.6}
      & 50.5 & \textbf{56.5}
      & 49.7 & \textbf{56.3}
\\
Hatebase
      & 82.3 & \textbf{86.1}
      & 84.1 & \textbf{87.9}
      & 82.5 & \textbf{83.6}
      &
      & 46.6 & \textbf{51.5}
      & 49.1 & \textbf{53.7}
      & 47.7 & \textbf{51.9}
\\
Civil Comments
      & 83.8 & \textbf{85.4} 
      & 82.8 & \textbf{84.7} 
      & 83.7 & \textbf{85.4}
      &
      & \textbf{50.8} & 50.5
      & \textbf{51.0} & \textbf{51.0}
      & 50.9          & \textbf{51.3}
\\
    \bottomrule
    \end{tabular}
    \caption{\textbf{Held-out target model}. Accuracy for the detection and labeling tasks evaluated on a held-out target model; $f^D$ and~$f^L$ are trained on attacks aimed at two out of the three possible target models~(BERT, RoBERTA, XLNet), and tested on attacks aimed at the remaining held-out target model. For example, the BERT columns represent accuracy for  models trained on attacks aimed at RoBERTA and XLNet, and evaluated on attacks aimed at BERT.} 
    \label{tab:heldout_target_model}
    \vskip -5mm
\end{table*}

\subsection{Generalizing to New Attack Method}
We also test the ability of our attack detection methods to identify new attacks by training them to detect all attacks except one, and then measuring the accuracy of the ``any attack'' detector on separating a new, previously unseen attack from clean data. We chose three attacks from TextAttack and OpenAttack: Pr., FG and VIP*. We isolate them and detect one at a time against RoBERTa. 
We compare the performance of L and T models with TLC features; average accuracy for each dataset is presented in Table~\ref{tab:heldout_attack}. Our detection methods show good detection accuracy on heldout attacks (82\% acc. for L and 86\% acc. for T), with T holding a slight advantage over L in the majority of cases.

\begin{table}[t]
\small
\setlength\tabcolsep{3pt}
\center
\begin{tabular}{l ccc ccc}
\toprule
  & \multicolumn{2}{c}{\textbf{Pr.}} & \multicolumn{2}{c}{\textbf{FG}} & \multicolumn{2}{c}{\textbf{VIP*}} \\
\cmidrule(lr){2-3}
\cmidrule(lr){4-5}
\cmidrule(lr){6-7}
\textbf{Dataset} &
 L\scriptsize-TLC & T\scriptsize-TLC &
 L\scriptsize-TLC & T\scriptsize-TLC &
 L\scriptsize-TLC & T\scriptsize-TLC \\
\midrule
Clim. Change
     & 79.1          & \textbf{80.7}
     & 80.7          & \textbf{84.6}
     & \textbf{85.7} & 82.9
\\
IMDB
      & \textbf{75.2} & 69.3
      & \textbf{89.5} & 82.6
      & 66.4          & \textbf{87.4}
\\
SST-2
      & 75.3 & \textbf{86.7}
      & 79.7 & \textbf{92.7}
      & 74.2 & \textbf{88.2}
\\
Wikipedia
      & 84.9          & \textbf{89.0}
      & 84.0          & \textbf{91.8} 
      & \textbf{88.9} & 87.9
\\
Hatebase
      & 93.3          & \textbf{98.3} 
      & 92.7          & \textbf{98.0} 
      & \textbf{97.0} & 95.2
\\
Civ. Comm.
      & 75.6          & \textbf{80.6} 
      & 75.0          & \textbf{81.5} 
      & \textbf{75.2} & 66.1
\\
    \bottomrule
    \end{tabular}
    \caption{\textbf{Held-out attacks}. Balanced detection accuracy of clean vs.\ each held-out attack. Each model is trained to separate clean instances from all attacks (except held-out ones) and tested on separating clean instances from each held-out attack.}
    \label{tab:heldout_attack}
        \vskip -5mm

\end{table}

\subsection{Analyzing Feature Importances}
\label{sec:analyzing_feature_importances}
 
We use the following as the contribution of each feature (or feature set) to predictions of each class:
 \begin{equation}\label{eu_eqn}
    I(A, c) = \frac{1}{m}\sum_{j=1}^{m}|x^{(j)}_{A} \cdot w^{(c)}_{A}|,
 \end{equation}
in which $c$ is a class (an attack or \emph{clean}),
 $m$ is the number of test instances, and $x_{A}$ and $w_{A}$ denote the segment of the instance and weight vectors associated with feature or feature set $A$, respectively.
 
\begin{figure}[tb]
    \centering
    \includegraphics[width=\linewidth]{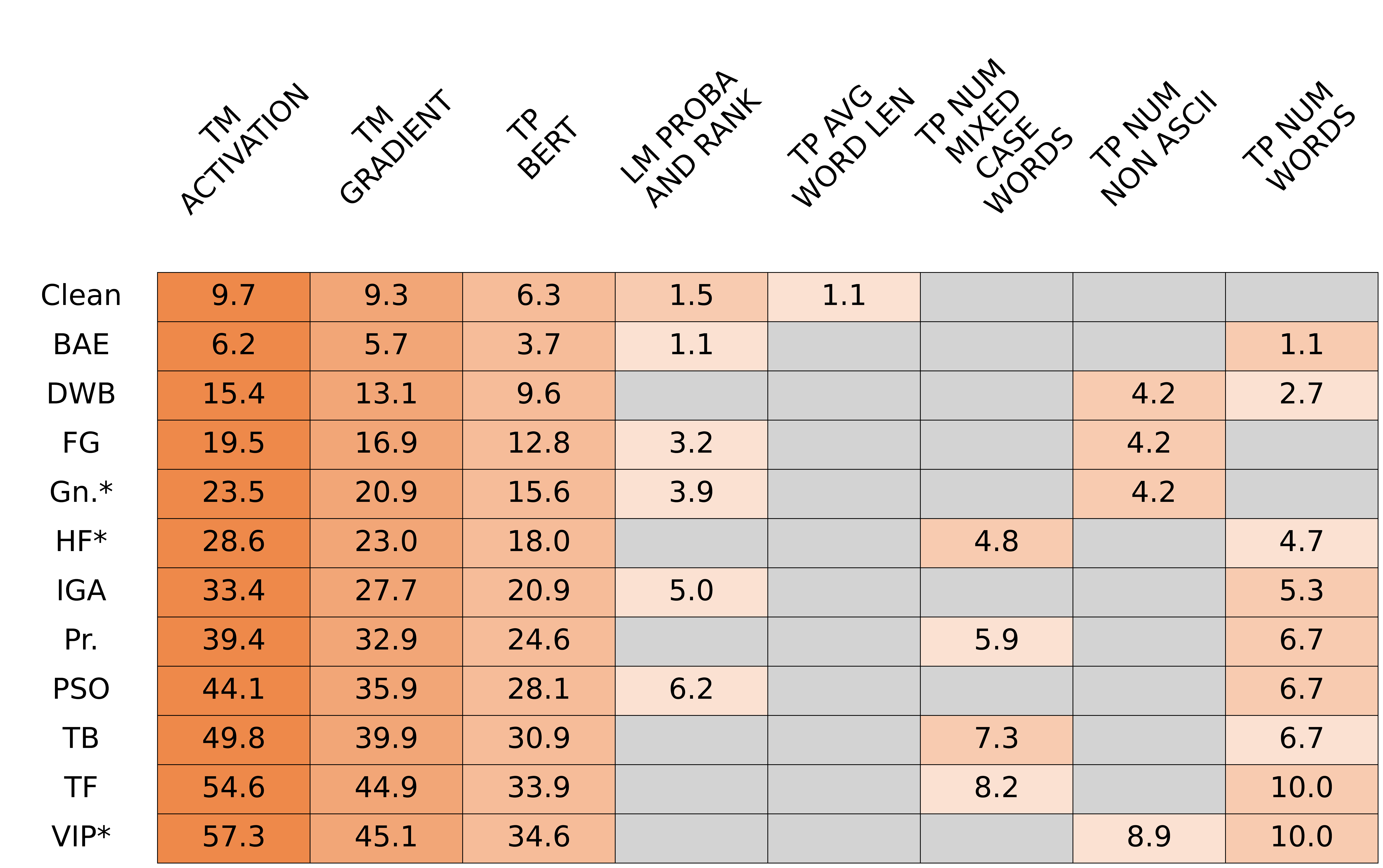}
    \caption{Top 5 Features (and their contributions, see Eq.~\ref{eu_eqn}) for each attack method for the L-TLC~(RoBERTa) labeling model on the Hatebase dataset. Gray indicates feature is not in the top 5. 
    }
    \label{fig:contrib_1}
    \vskip -5mm
\end{figure}

Figure~\ref{fig:contrib_1} shows the contribution of the top six features from the L-TLC~(RoBERTa) attack labeling model for Hatebase~(see Table~\ref{tab:all_features} in the Appendix for all feature contributions). Target \textbf{C}lassifier properties (activations and gradients) and BERT representations show consistent heavy impact across all attacks and clean data. This is consistent with our findings described in \S\ref{detection} where L and T models with TLC features outperform other feature ablations in nearly all cases. 
 
\section{Related Work}
\label{sec:related_work}

We now review prior work on making models robust to adversarial attacks, detecting adversarial examples, and frameworks and datasets for training and evaluating the robustness of NLP models.

\paragraph{Building Robust Models}
Much of the work on defending against textual adversarial attacks consists of building more robust predictive models through some form of adversarial training.
\citet{liu2020robust} and \citet{tan2020mind} augment their datasets with adversarial examples, \citet{dong2021towards,wang2020infobert, Ivgi2021AchievingMR, yoo-qi-2021-towards-improving, Wang2021AdversarialTW, miyato2017adversarial, Zhu2020FreeLB} train models on adversarial examples generated in an online fashion, and~\citet{dinan2019build} uses humans-in-the-loop to generate high-quality adversarial examples. \citet{malykh2019robust,jones2020robust,liu2020joint} encode the input to leverage the fact that perturbations will be close to the original unperturbed input.

Other methods of creating robust models have been proposed as well.
\citet{jiang2020smart} and~\citet{li2016robust_representations} modify the training objective to control the complexity of the model.~\citet{jones2020robust} introduce a robust encoding layer and~\citet{ye2020safer} develop a randomized smoothing method; both provide some degree of guaranteed robustness even for large models like BERT.~\citet{huang2019ibp} and~\citet{jia2019certified} both use interval bound propagation to train models that have guaranteed robustness to word substitutions.
\citet{shi2020verification} develop a new robustness verification algorithm that applies to transformers and produces tighter bounds than naive interval bound propagation.

\paragraph{Perturbation Detection}

In contrast to building more robust models, \citet{zhou2019learning} train a model using contextualized BERT features to detect token/word perturbations and attempt to fix perturbed instances by replacing any perturbed word/token with a suitable replacement from a learned input space.
Similarly,~\citet{xie-etal-2021-models} propose categorizing adversarial attacks using BERT sentence embeddings and target model activations.
\citet{pruthi2019combating} use a word-recognition model to combat word-mispelling attacks, and~\citet{mozes-etal-2021-frequency} compare inputs before and after replacing infrequent words with more frequent words to detect word-substitution attacks.
\citet{le2020detecting} inject multiple trapdoors into textual classifiers, baiting attackers with local optima to detect universal triggers~\cite{wallace2019universal}. \citet{hovy2016enemy} train a logistic regression model to detect fake reviews using word n-grams 
plus review meta-information for improved performance.
 \citet{Li2021ContextualizedPF} show that adversarial examples often have high perplexity, as measured by a language model.

While prior work on labeling text attacks is limited, there is analogous work on attribution of GAN-generated images. \citet{yu19attributing} find that GANs generate stable ``fingerprints'' which can be used for fine-grained attribution. 
\citet{albright19source} determine if a given image was generated by a specific GAN by reversing the generation process. 
Since text perturbations are typically much sparser than pixel perturbations, attack attribution is much harder with short text.

\paragraph{Robustness Evaluation Frameworks/Datasets}
Evaluation frameworks such as Robustness Gym~\cite{goel2021robustness_gym} and TextFLINT~\cite{gui2021textflint} allow users to measure the performance of their own models on a variety of text transformations and adversaries. TextFLINT also makes available a dataset of 67,000 transformed text samples for training. Adversarial GLUE~\cite{wang2021adversarial} is a multi-task robustness benchmark that was created by applying 14 textual adversarial attack methods to GLUE tasks. Dynabench~\cite{keila2021dynabench} is a related framework for evaluating and training NLP models on adversarial examples created by human adversaries.

Of these, TCAB is most similar to Adversarial GLUE. However, TCAB was designed for a different task --- attack identification rather than robustness evaluation. TCAB is also much larger than Adversarial GLUE~(1.5 million fully automated attacks vs.~5,000 human-verified attacks), focuses only on classification domains, and includes multiple classification domain datasets.

\section{Conclusions and Future Work}

Unlike general robust training methods, attack identification attempts to understand existing ``low-effort'' attacks. 
Using our new TCAB attack identification benchmark, we find that attack detection works well and generalizes to unseen attacks. We achieve positive results for attack labeling as well, though similarities among different attacks make it difficult to determine the exact attack used from a single example. Real-world settings may actually be easier --- given a group of related adversarial examples, such as multiple abusive messages from a single account, there can be more clues to determine which attack was used to create the group.

However, this is just a first step. As new attacks are developed, we hope to expand TCAB with more attacks, as well as more domain datasets. We also expect that better detection and labeling accuracies are possible with more complex classifiers. 
Furthermore, while the confusion matrix reveals some relationships among the attacks, we expect that more structure could be discovered, especially as the number of attacks grows. Unknown attacks could be classified within this taxonomy based on their relationship to previously seen attacks. 
Another challenge is finding better features that generalize across different domains. The attacks themselves are not domain-specific, but we currently need to train on attacks for a specific domain dataset (e.g., IMDB) in order to label attacks in that domain. 
Finally, the true test of these methods is their effectiveness ``in the wild'' on attacks by actual adversaries. If attack identification leads to attacker understanding, then it is a critical piece of a comprehensive defense of NLP systems.

\clearpage
\section{Broader Impacts}

Attack identification can be used to fight spam, abuse, fake news, and more; however, adversarial attacks can also be used for evading surveillance and maintaining privacy. The same tools used to learn more about spammers could also reveal information about dissidents. Since adversarial attacks have dual use, defenses against them have dual use as well. See \citet{albert20politics} for a more extensive discussion of the political dimensions of adversarial machine learning, much of which applies to attack labeling as well.

\section*{Acknowledgements}

This work was supported by the Defense Advanced Research Projects Agency (DARPA), agreement number HR00112090135. This work benefited from access to the University of Oregon high performance computer, Talapas.

\bibliography{new_ref}
\bibliographystyle{styles/acl_natbib}

\onecolumn
\appendix

\section{Algorithmic Details}

\subsection{Samplewise Properties}
\label{sec_appendix:samplewise_properties}

This section provides a detailed description of each feature for the text, language model, and target model properties.

\subsubsection*{Text Properties}

\begin{itemize}[nosep]
    \item \textbf{BERT features}: BERT embedding representation of the input sequence.

    \item \textbf{No. chars.}: Number of characters.

    \item \textbf{No. alpha chars.}: Number of alphabet characters (a-z).

    \item \textbf{No. digit chars.}: Number of digit characters (0-9).

    \item \textbf{No. punctuation.}: Number of punctuation mark characters (“?”, “!”, etc.)

    \item \textbf{No. multi. spaces.}: Number of times multiple spaces appear between words.

    \item \textbf{No. words.}: Number of words.

    \item \textbf{Avg. word len.}: Mean, variance, and quantiles (25\%, 50\%, and 75\%) of the number of characters per word for different regions of the input (first 25\%, middle 50\%, last 25\%, entire input).

    \item \textbf{No. non-ascii.}: Number of non-ascii characters.

    \item \textbf{Cased letters.}: Number of uppercase letters, number of lowercase letters, fraction of uppercase letters, and fraction of lowercase letters.

    \item \textbf{Is first word lowercase.}: True if the first character of the first word is lowercase, otherwise False.

    \item \textbf{No. mixed-case words.}: Number of words that contain both upper and lowercase letters (not including the first letter of each word).

    \item \textbf{No. single lowercase letters.}: Number of single-letter-lowercase words (excluding “a” and “i”).

    \item \textbf{No. lowercase after punctuation.}: Number of words after a punctuation that begin with a lowercase letter.

    \item \textbf{No. cased word switches.}: Number of times words switch from all uppercase to all lowercase and vice versa (e.g. THIS IS a sentence THAT contains 3 switches).
\end{itemize}

\subsubsection*{Language Model Properties}

\begin{itemize}[nosep]
    \item \textbf{Probability and rank}: Mean, variance, and quantiles (25\%, 50\% and 75\%) of the token probabilities and ranks for different regions of the input (first 25\%, middle 50\%, last 25\%, entire input) using RoBERTa, a masked language model.

    \item \textbf{Perplexity}: Perplexity for different regions of the input (first 25\%, middle 50\%, last 25\%, entire input) using GPT-2, a causal language model.
\end{itemize}

\subsubsection*{Target Model Properties}

For the following properties, we assume the target model to be a RoBERTa text classifier.

\begin{itemize}[nosep]
    \item \textbf{Posterior.}: Output posteriors of the target model (softmax applied to logits).

    \item \textbf{Gradient.} Mean, variance, and quantiles (25\%, 50\%, and 75\%) of the gradients for each layer of the target model given different regions of the input (first 25\%, middle 50\%, last 25\%, entire input).

    \item \textbf{Activation.}: Mean, variance, and quantiles (25\%, 50\%, and 75\%) of the node activations for each layer of the target model given different regions of the input (first 25\%, middle 50\%, last 25\%, entire input).

    \item \textbf{Saliency.}: Mean, variance, and quantiles (25\%, 50\%, and 75\%) of the saliency values (gradients of the target model with respect to the input tokens) given the input.
\end{itemize}

\newpage
\section{Experiment Details}

Experiments are run on a TITAN RTX GPU with 24GB of memory and an Intel(R) Xeon(R) CPU E5-2690 v4 @ 2.6GHz with 60GB of memory. Experiments are run using Python 3.8. Source code for generating the attack dataset and all experiments will be made public upon publication.

\subsection{Domain Datasets}

\begin{itemize}[nosep]
    \item Climate Change\footnote{\url{https://www.kaggle.com/edqian/twitter-climate-change-sentiment-dataset}} consists of 62,356 tweets from Twitter pertaining to climate change. The collection of this data was funded by a Canada Foundation for Innovation JELF Grant to Chris Bauch, University of Waterloo. The goal is to determine if the text has a negative, neutral, or positive sentiment.

    \item IMDB consists of 50,000 highly polar movie reviews collected from \url{imdb.com} curated by~\citet{maas-EtAl:2011:ACL-HLT2011}. The goal is to determine if the text has a negative or positive sentiment.

    \item SST-2~\cite{socher2013recursive} contains 68,221 phrases with fine-grained sentiment labels in the parse trees of 11,855 sentences from movie reviews. The goal is to determine if the text has a negative or positive sentiment.

    \item Wikipedia (Talk Pages) consists of 159,686 comments~(9.6\% toxic) from Wikipedia editorial talk pages. The data was curated by~\citet{wulczyn2017ex} and made readily available by~\citet{dixon2018measuring}. The goal is to distinguish between non-toxic and toxic comments.

    \item Hatebase~\cite{davidson2017automated} consists of 24,783~(83.2\% toxic) tweets from Twitter collected via searching for tweets containing words from the lexicon provided by \url{hatebase.org}. The goal is to distinguish between non-toxic and toxic comments.

    \item Civil Comments\footnote{\url{https://www.kaggle.com/c/jigsaw-unintended-bias-in-toxicity-classification}} consists of 1,804,874~(8\% toxic) messages collected from the platform Civil Comments. The goal is to distinguish between non-toxic and toxic comments.
\end{itemize}

For datasets without a predefined split, we use an 80/10/10 train/validation/test split. 

\newpage
\subsection{Target Models}

We use the popular HuggingFace transformers library\footnote{\url{https://huggingface.co/}} to fine-tune three transformer-based models designed for text classification~(BERT~\cite{devlin2019bert}, RoBERTa~\cite{liu2019roberta}, and XLNet~\cite{yang2019xlnet}) on each domain dataset.
Table~\ref{tab:target_model_training_hyperparameters} shows the training parameters used to fine-tune/train each model. We use cross entropy as the loss function and Adam as the optimizer to train all models.

\begin{table*}[h]
\small
\setlength\tabcolsep{1.8pt}
\center
\begin{tabular}{llrrrrrr}
\toprule
\textbf{Dataset} & \textbf{Model} & \textbf{Max. len.} & \textbf{Learning rate} & \textbf{Batch size} & \textbf{No. epochs} & \textbf{Decay} & \textbf{Max. norm} \\
\midrule
\multirow{3}{*}{Climate Change}
    & BERT    & 250 & $1e^{-5}$ & 64 & 15 & 0 & 1.0 \\
    & RoBERTa & 250 & $1e^{-5}$ & 64 & 15 & 0 & 1.0 \\
    & XLNet   & 250 & $4e^{-5}$ & 64 & 15 & 0 & 1.0 \\
\midrule
\multirow{3}{*}{IMDB}
    & BERT    & 128 & $4e^{-5}$ & 64 & 5  & 0 & 1.0 \\
    & RoBERTa & 128 & $1e^{-6}$ & 64 & 10 & 0 & 1.0 \\
    & XLNet   & 128 & $4e^{-5}$ & 64 & 5  & 0 & 1.0 \\
\midrule
\multirow{3}{*}{SST-2}
    & BERT    & 128 & $1e^{-5}$ & 32 & 5 & 0 & 1.0 \\
    & RoBERTa & 128 & $1e^{-5}$ & 32 & 5 & 0 & 1.0 \\
    & XLNet   & 128 & $1e^{-5}$ & 32 & 5 & 0 & 1.0 \\
\midrule
\multirow{3}{*}{Wikipedia}
    & BERT    & 250 & $1e^{-6}$ & 32 & 10 & 0 & 1.0 \\
    & RoBERTa & 250 & $1e^{-6}$ & 32 & 10 & 0 & 1.0 \\
    & XLNet   & 250 & $1e^{-6}$ & 16 & 10 & 0 & 1.0 \\
\midrule
\multirow{3}{*}{Hatebase}
    & BERT    & 250 & $1e^{-6}$ & 32 & 50 & 0 & 1.0 \\
    & RoBERTa & 250 & $1e^{-6}$ & 32 & 50 & 0 & 1.0 \\
    & XLNet   & 128 & $1e^{-6}$ & 16 & 50 & 0 & 1.0 \\
\midrule
\multirow{3}{*}{Civil Comments}
    & BERT    & 250 & $1e^{-6}$ & 32 & 10 & 0 & 1.0 \\
    & RoBERTa & 250 & $1e^{-6}$ & 32 & 10 & 0 & 1.0 \\
    & XLNet   & 128 & $1e^{-6}$ & 16 & 10 & 0 & 1.0 \\
\bottomrule
\end{tabular}
\caption{Training parameters used to fine-tune/train each target model. Max. len. is the maximum number of tokens fed into each model; Decay denotes the weight decay of the model.}
\label{tab:target_model_training_hyperparameters}
\end{table*}

\newpage
\subsection{Detection and Attack Identification}
Table~\ref{tab:clean_vs_all} shows accuracy scores for the detection task. Note that the best accuracy score always occurs when all of three feature categories (TLC) are used. The boosted trees outperform logistic regression in most cases. Also note that in the cases when the target model does not match the model that was used to create the ``classifier'' features (i.e. the BRT and XLN columns), accuracy is only slightly worse than when the target model \textit{does} match the model used to create them (i.e. the RBT column).

\begin{table*}[h]
\small
\setlength\tabcolsep{2.5pt}
\center
\begin{tabular}{l rrr rrr rrr rrr rrr rrr}
\toprule
 \multirow{2}{*}{\bf Model}
 & \multicolumn{3}{c}{\textbf{Climate Change}} & \multicolumn{3}{c}{\textbf{IMDB}} & \multicolumn{3}{c}{\textbf{SST-2}} & \multicolumn{3}{c}{\textbf{Wikipedia}} & \multicolumn{3}{c}{\textbf{Hatebase}} & \multicolumn{3}{c}{\textbf{Civil Comments}} \\
\cmidrule(lr){2-4}
\cmidrule(lr){5-7}
\cmidrule(lr){8-10}
\cmidrule(lr){11-13}
\cmidrule(lr){14-16}
\cmidrule(lr){17-19}
 & BRT & RBT & XLN & BRT & RBT & XLN & BRT & RBT & XLN & BRT & RBT & XLN & BRT & RBT & XLN & BRT & RBT & XLN\\
\midrule
L       & 73.8 & 76.6 & 73.1 & 
    76.6 & 83.2 & 80.5 & 
    75.6 & 75.5 & 79.7 &  
    82.6 & 84.1 & 84.5 &    
    77.4 & 76.5 & 79.6 & 
    75.6 & 75.6 & 75.1 \\
L-T     & 78.0 & 80.8 & 76.0 &    
    84.3 & 90.4 & 85.6 &    
    86.0 & 86.1 & 84.8 & 
    83.4 & 85.1 & 84.5 &    
    82.9 & 82.1 & 79.8 & 
    76.9 & 77.2 & 77.1 \\
L-TL    & 82.2 & 85.3 & 81.2 &    
    85.5 & 91.0 & 87.3 &    
    86.5 & 87.0 & 86.0 & 
    83.2 & 85.3 & 85.3 &    
    82.6 & 82.0 & 79.9 & 
    79.4 & 79.7 & 80.3 \\
L-TLC   & \textbf{85.1} & \textbf{88.6} & \textbf{84.1} &    
    \textbf{89.5} & \textbf{95.2} & \textbf{90.6} &    
    \textbf{86.9} & \textbf{87.8} & \textbf{87.7} &    
    \textbf{87.9} & \textbf{90.1} & \textbf{87.8} &    
    \textbf{87.3} & \textbf{91.2} & \textbf{86.2} & 
    \textbf{84.3} & \textbf{84.2} & \textbf{84.4} \\
\addlinespace
T       & 69.6 & 71.6 & 69.3 &    
    70.6 & 76.6 & 74.5 &    
    72.8 & 73.2 & 76.7 & 
    82.5 & 82.3 & 83.6 &    
    69.5 & 69.7 & 75.9 & 
    74.0 & 73.3 & 72.3 \\
T-T     & 75.4 & 78.0 & 73.5 &    
    86.2 & 91.5 & 83.9 &    
    86.1 & 84.4 & 81.9 & 
    82.6 & 84.0 & 84.1 &    
    79.5 & 79.1 & 77.3 & 
    77.3 & 75.4 & 74.0 \\
T-TL    & 82.0 & 84.5 & 80.9 &    
    88.4 & 92.9 & 88.2 &    
    87.0 & 86.5 & 85.2 & 
    84.4 & 86.0 & 85.4 &    
    80.3 & 82.4 & 79.3 & 
    79.7 & 79.0 & 77.6 \\
T-TLC   & \textbf{84.2} & \textbf{88.2} & \textbf{83.1} &    
    \textbf{90.6} & \textbf{97.2} & \textbf{91.0} &    
    \textbf{88.2} & \textbf{89.5} & \textbf{87.5 }& 
    \textbf{90.9} & \textbf{93.3} & \textbf{91.1} &    
    \textbf{89.5} & \textbf{93.0} & \textbf{85.0} & 
    \textbf{84.5} & \textbf{84.4} & \textbf{84.6} \\
\bottomrule
\end{tabular}
\caption{Accuracy for clean vs. all attacks (binary). Attacked samples and clean samples are balanced 50/50. The best number in each column for each group (L vs. T) is bolded.}
\label{tab:clean_vs_all}
\end{table*}

Table \ref{tab:mulitclass_with_clean} shows accuracy scores for the multiclass attack labeling task. Note that the best accuracy score almost always occurs when all of three feature categories (TLC) are used. As with detection, boosted trees usually outperforms logistic regression. And also as with the detection figures, note that in the cases when the target model does not match the model that was used to create the ``classifier'' features (i.e. the BRT and XLN columns), accuracy is only slightly worse than when the target model \textit{does} match the model used to create them (i.e. the RBT column).

\begin{table*}[h]
\small
\setlength\tabcolsep{2.5pt}
\center
\begin{tabular}{l rrr rrr rrr rrr rrr rrr}
\toprule
\multirow{2}{*}{\bf Model}
 & \multicolumn{3}{c}{\textbf{Climate Change}} & \multicolumn{3}{c}{\textbf{IMDB}} & \multicolumn{3}{c}{\textbf{SST-2}} & \multicolumn{3}{c}{\textbf{Wikipedia}} & \multicolumn{3}{c}{\textbf{Hatebase}} & \multicolumn{3}{c}{\textbf{Civil Comments}} \\
\cmidrule(lr){2-4}
\cmidrule(lr){5-7}
\cmidrule(lr){8-10}
\cmidrule(lr){11-13}
\cmidrule(lr){14-16}
\cmidrule(lr){17-19}
& BRT & RBT & XLN & BRT & RBT & XLN & BRT & RBT & XLN & BRT & RBT & XLN & BRT & RBT & XLN & BRT & RBT & XLN \\
\midrule
L   & 37.9 & 48.5 & 45.8 &    
    38.7 & 39.6 & 45.5 &    
    38.8 & 39.0 & 44.5 &    
    32.7 & 34.5 & 34.0 &    
    38.0 & 38.7 & 35.4 & 
    29.6 & 30.4 & 38.3 \\
L-T & 53.9 & 64.1 & 59.8 &    
    55.0 & 55.5 & 59.9 &    
    56.2 & 56.9 & 57.8 &    
    43.4 & \textbf{44.2} & 45.1 &    
    50.5 & 49.8 & \textbf{43.4} & 
    38.0 & 38.8 & 46.1 \\
L-TL    & 55.6 & 65.5 & 61.5 &    
    57.2 & 57.8 & 62.8 &    
    56.6 & 57.3 & 58.7 &    
    43.3 & 44.1 & \textbf{46.3} &    
    49.6 & 50.6 & 43.0 & 
    39.2 & 40.0 & 47.4 \\
L-TLC   & \textbf{58.9} & \textbf{68.5} & \textbf{64.8} &    
    \textbf{60.8} & \textbf{62.0} & \textbf{65.6} &    
    \textbf{58.9} & \textbf{61.8} & \textbf{61.4} &    
    \textbf{44.3} & 43.3 & 44.8 &    
    \textbf{51.5} & \textbf{52.5} & 43.1 & 
    \textbf{45.2} & \textbf{45.3} & \textbf{51.9} \\
\addlinespace
T   & 32.4 & 38.9 & 37.2 &    
    27.4 & 28.5 & 33.5 &    
    31.7 & 32.0 & 37.1 &    
    26.1 & 24.8 & 24.8 &    
    28.5 & 32.1 & 32.1 & 
    28.0 & 22.8 & 23.8 \\
T-T & 52.6 & 63.0 & 57.9 &    
    63.8 & 61.2 & 64.3 &    
    53.4 & 53.8 & 53.9 &    
    45.8 & 47.5 & 47.5 &    
    48.4 & 48.1 & 38.6 & 
    41.3 & 37.1 & 37.4 \\
T-TL    & 55.4 & 65.6 & 61.5 &    
    65.8 & 62.5 & 66.4 &    
    56.6 & 56.6 & 57.1 &    
    48.0 & 50.0 & 50.0 &    
    \textbf{49.4} & 49.1 & 41.9 & 
    44.6 & 40.1 & 40.3 \\
T-TLC   & \textbf{58.4} & \textbf{69.4} & \textbf{64.5} &    
    \textbf{69.7} & \textbf{69.3} & \textbf{70.9} &    
    \textbf{60.9} & \textbf{62.8} & \textbf{61.7} &    
    \textbf{50.5} & \textbf{51.5} & \textbf{52.2} &    
    47.9 & \textbf{55.1} & \textbf{44.7} & 
    \textbf{44.9} & \textbf{45.4} & \textbf{50.7} \\
\bottomrule
\end{tabular}
\caption{Accuracy for the attack labeling task.  
Baseline accuracy is approximately 1/12 = 8.33\%.  The best number in each column for each group (L vs. T) is bolded.}
\label{tab:mulitclass_with_clean}
\end{table*}

\newpage
\subsection{Attack Methods}
In Table \ref{tab:attack_methods}, all of the attack methods used to create TCAB are listed along with the toolchain each attack method belongs to and the access level, linguistic constraints, and perturbation level each attack method has.
\begin{table}[h]
\center
\begin{tabular}{l l l l l}
\toprule
\textbf{Attack method} & \textbf{Toolchain} & \textbf{Access level} & \textbf{Linguistic constraints} & \textbf{Perturbation level} \\
\midrule
BAE & TextAttack & Black box & Yes & Word \\
DeepWordBug & TextAttack & Gray box  & No & Char \\
FasterGenetic & TextAttack & Gray box  & Yes & Word \\
Genetic & OpenAttack & Gray box  & Yes & Word \\
HotFlip & OpenAttack & White box & Yes & Char \\
IGA & TextAttack & Gray box  & Yes & Word \\
Pruthi & TextAttack & Gray box  & No & Word \\
PSO & TextAttack & Black box & Yes & Word \\
TextBugger & TextAttack & Black box & Yes & Char \\
TextBugger & OpenAttack & White box & Yes & Word + Char \\
TextFooler & TextAttack & Black box & Yes & Word \\
VIPER & OpenAttack & Black box & No & Char \\
\bottomrule
\end{tabular}
\caption{The 12 attack methods used to create TCAB.}
\label{tab:attack_methods}
\end{table}

\newpage
\subsection{Attack Labeling: Confusion Matrix}
\label{appendix_sec:confusion_matrix}

\begin{figure}[h!]
\centering
\subcaptionbox{Climate Change}{\includegraphics[width=0.49\textwidth]{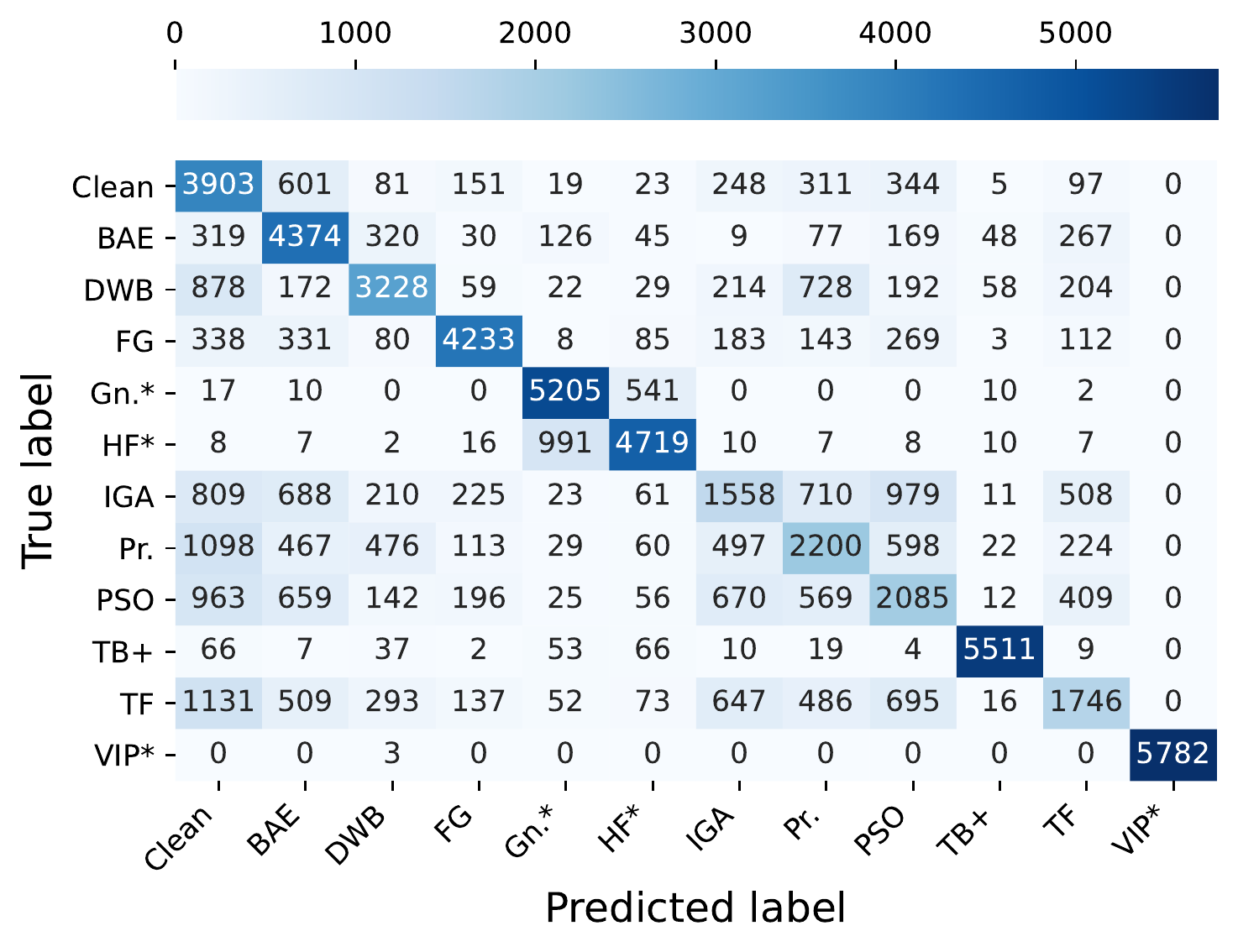}}
\hfill
\subcaptionbox{IMDB: IGAWang is not included since it has no successful attacks on the Wikipedia dataset.}{\includegraphics[width=0.49\textwidth]{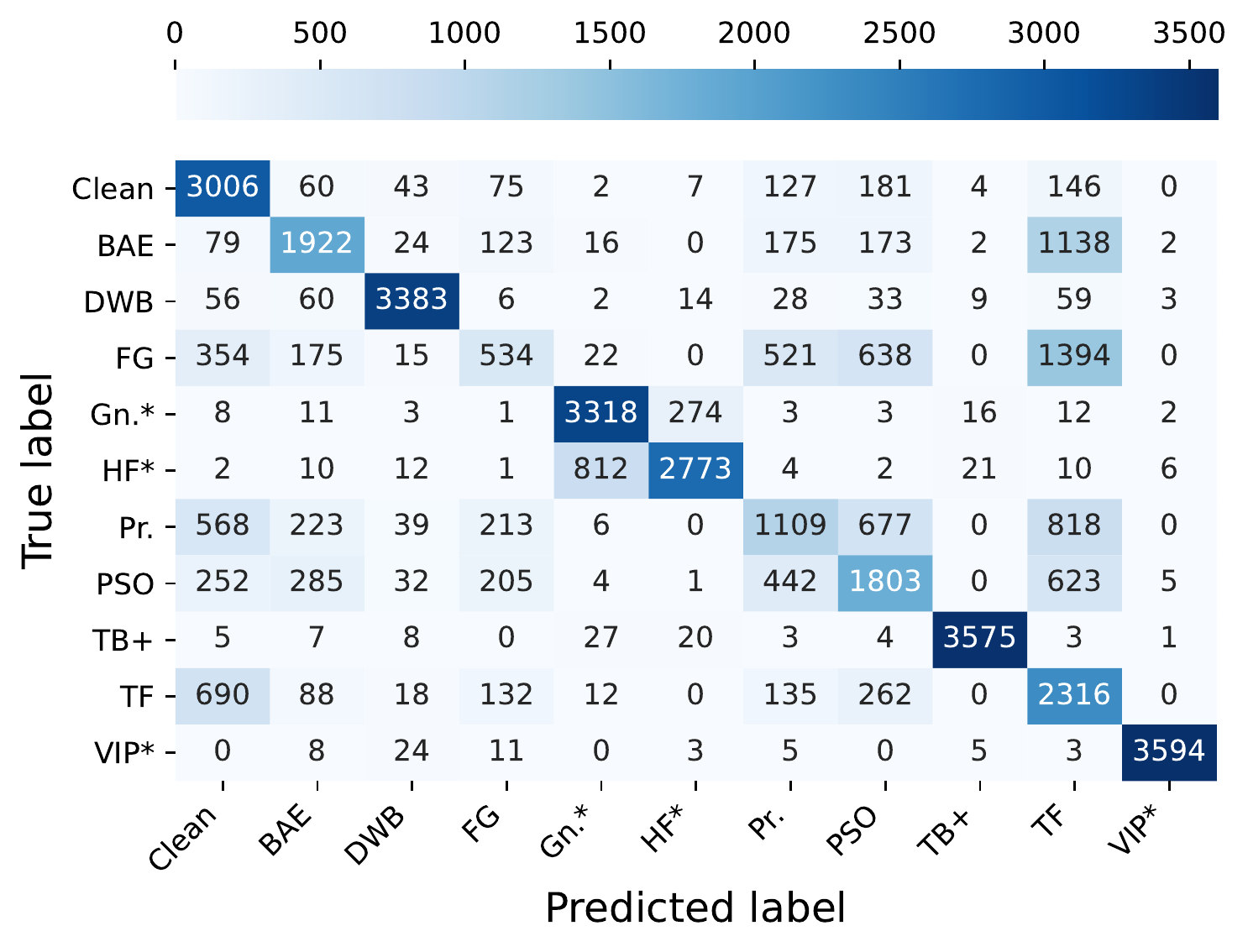}}
\hfill
\subcaptionbox{SST-2}{\includegraphics[width=0.49\textwidth]{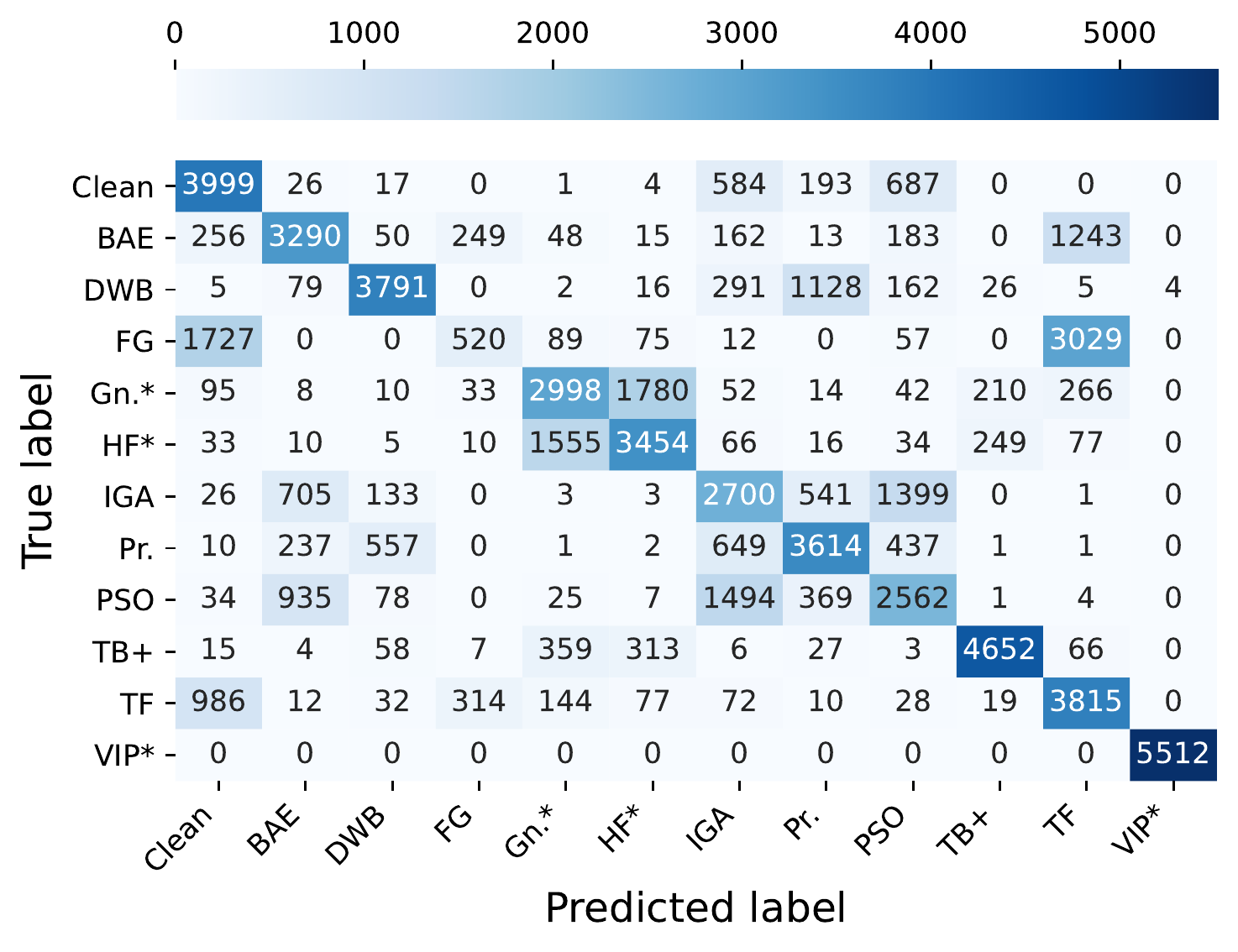}}
\hfill
\subcaptionbox{Wikipedia: IGAWang is not included since it has no successful attacks on the Wikipedia dataset.}{\includegraphics[width=0.49\textwidth]{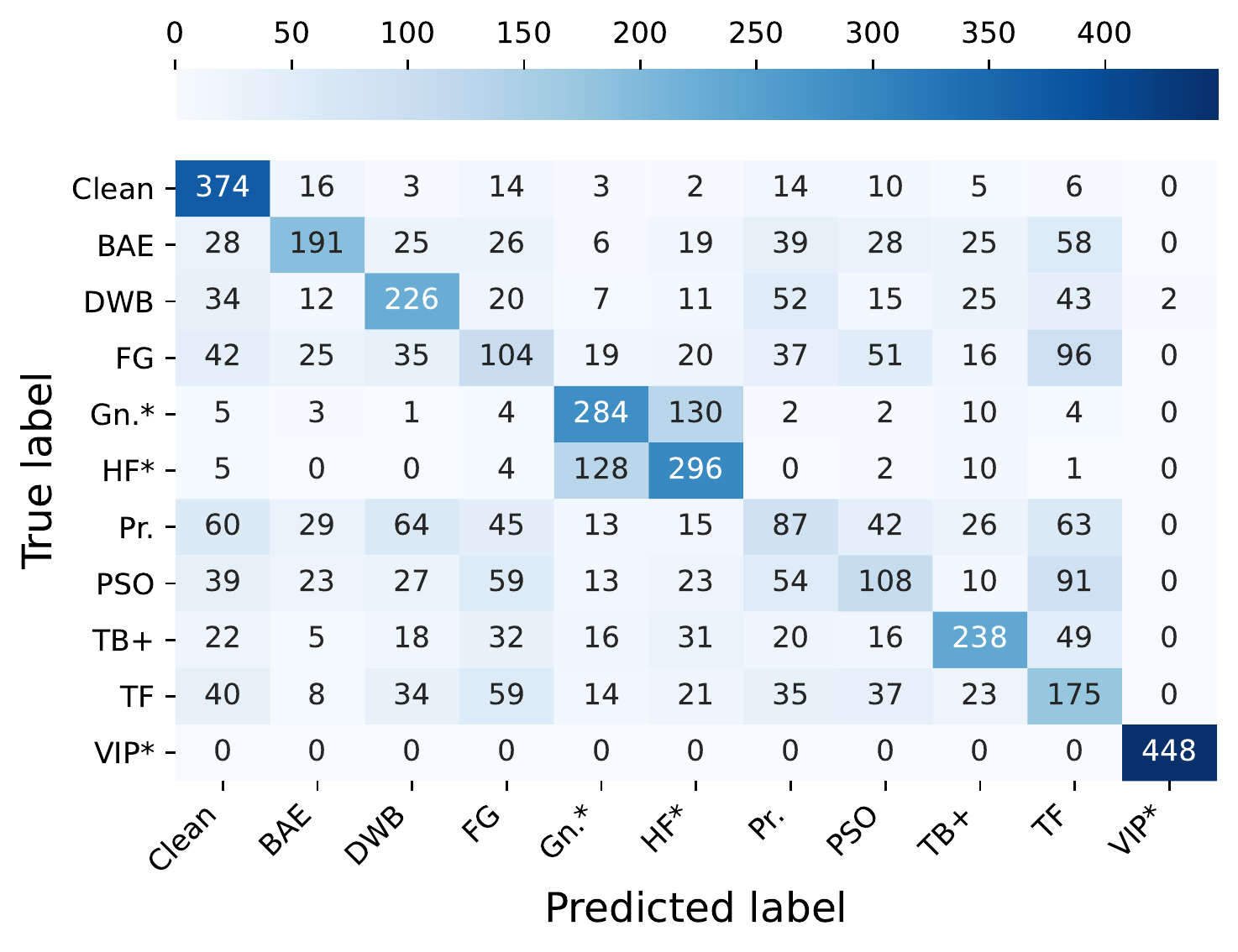}}
\hfill \\
\subcaptionbox{Hatebase}{\includegraphics[width=0.49\textwidth]{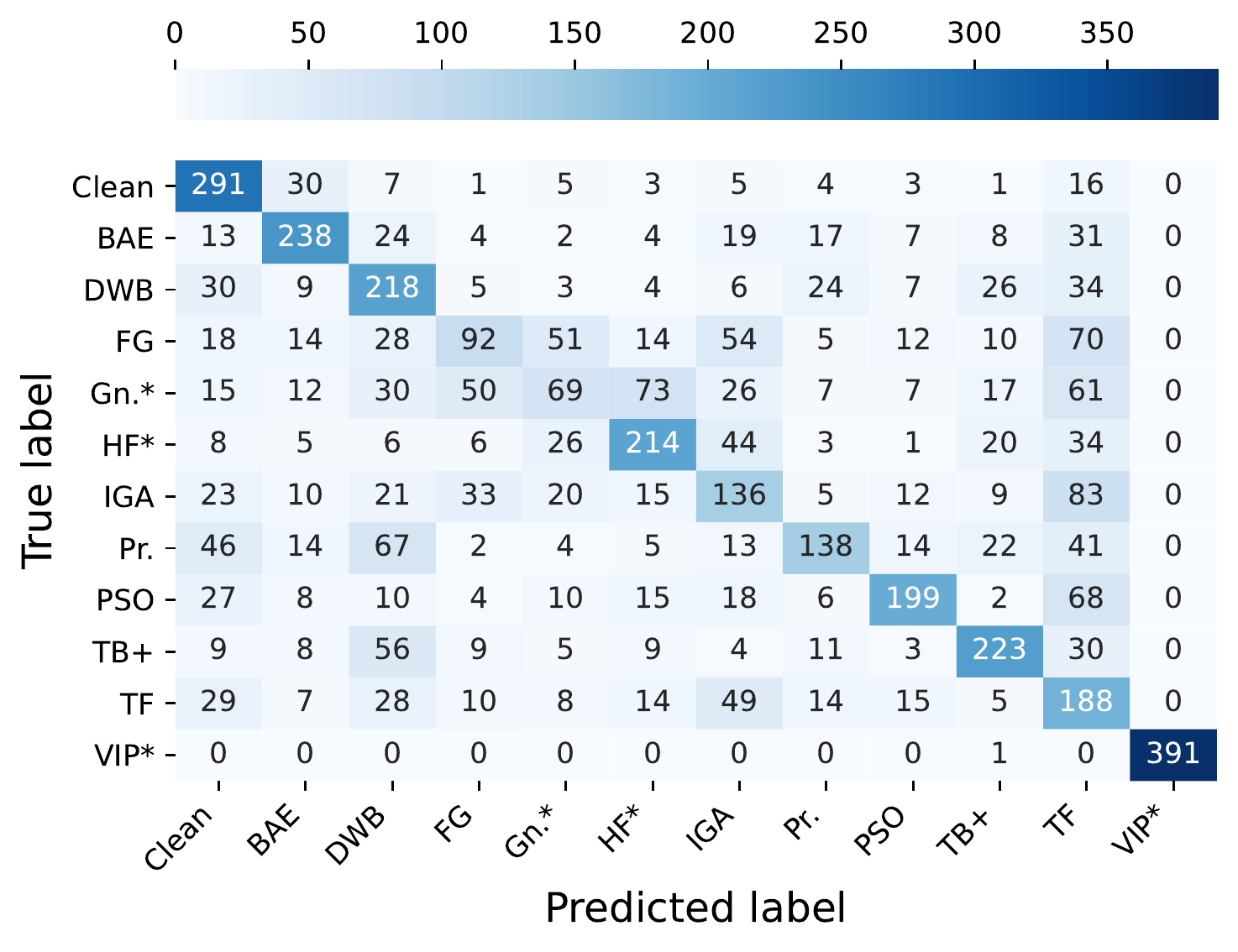}}
\hfill
\subcaptionbox{Civil Comments}{\includegraphics[width=0.49\textwidth]{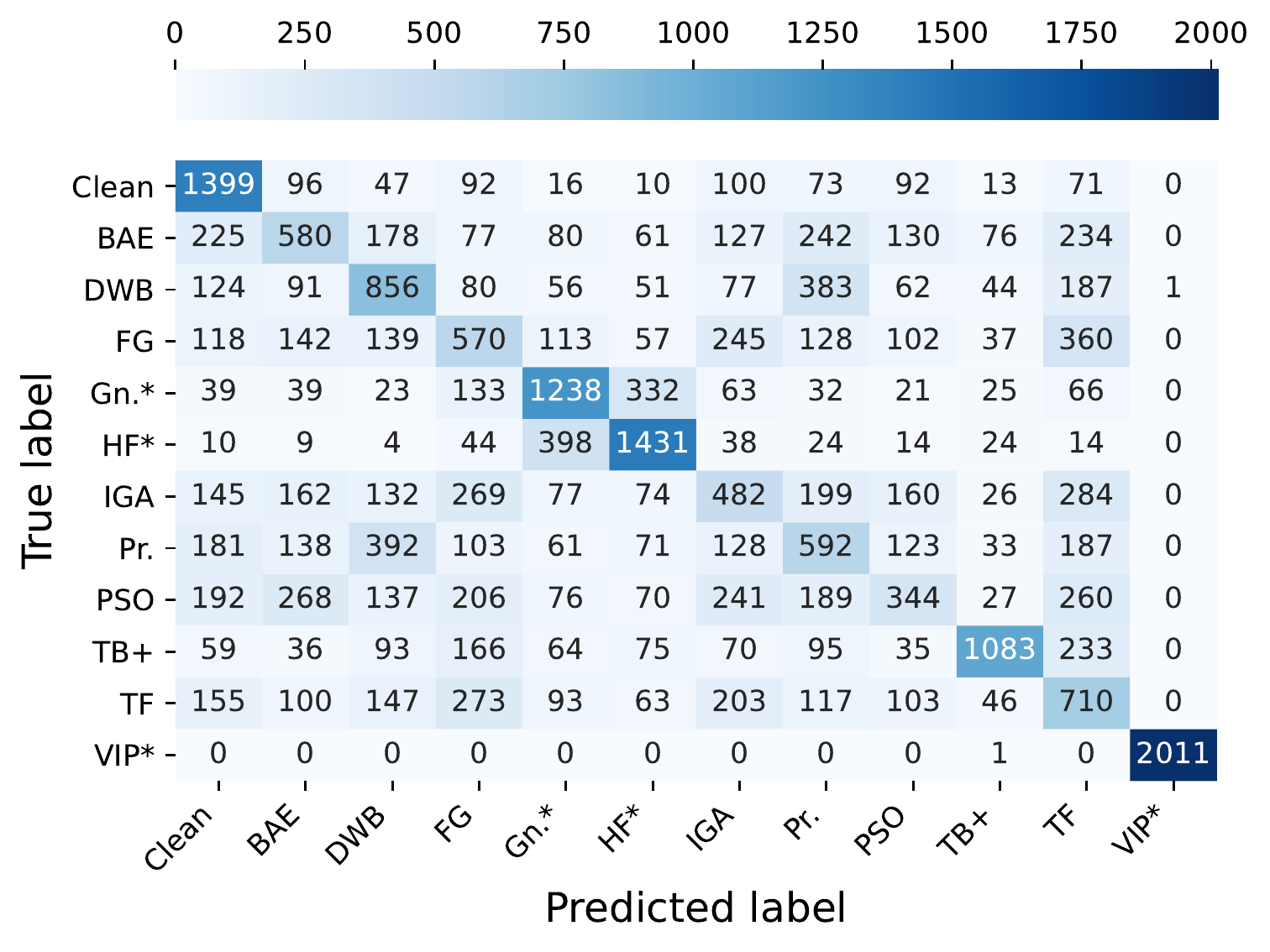}}
\caption{Confusion matrix results using the T-TLC model to label attacks, averaged over all target models.}
\label{fig:appendix_confusion_matrix}
\end{figure}

Figure~\ref{fig:appendix_confusion_matrix} shows the confusion matrix results for each domain dataset using T-TLC to label attacks, averaged over all target models.

\newpage
\subsection{Attack Samples}
\label{sec:attack_samples}

We present Tables \ref{tab:attack_samples_sst}, \ref{tab:attack_samples_wikipedia_1} and \ref{tab:attack_samples_wikipedia_2} with a few successful attack samples on the same pieces of text from the SST and the Wikipedia datasets predicted by XLNet model attacked by a variety of attackers respectively; and Table \ref{tab:attack_samples_sst_multi} with a few successful attack samples on different pieces of text from SST predicted by RoBERTa model and attacked by BAE.

\begin{table*}[h]
\small
\setlength\tabcolsep{3pt}
\center
\begin{tabular}{lm{9cm}cr}
\toprule
\textbf{Attack} & \textbf{Text} & \textbf{Label} & \textbf{Confidence}\\
\midrule
    Original
    & Part comedy, part drama, the movie winds up accomplishing neither in full, and leaves us feeling touched and amused by several moments and ideas, but nevertheless dissatisfied with the movie as a whole.
    & Negative 
    & 89.2\% \\
\midrule
    BAE
    & Part comedy, part drama, the movie winds up accomplishing neither in full, and leaves us feeling touched and amused by several moments and ideas, but nevertheless \textcolor{red}{satisfied} with the movie as a whole.
    & Positive 
    & 84.4\%\\
\midrule
    DeepWordBug
    &Part comedy, part drama, the movie wi\textcolor{red}{r}nds up accomplishing \textcolor{red}{en}ither in full, and leaves us feeling touched and amused by several moments and ideas, but nevertheless dissat\textcolor{red}{A}sfied with the movie as a whole.
    & Positive 
    & 60.1\%\\
\midrule
    FasterGenetic
    & Part comedy, part drama, the movie winds up accomplishing \textcolor{red}{or} in full, and leaves us feeling touched and amused by several moments and ideas, but \textcolor{red}{notwithstanding displeased} with the movie as a whole.  
    & Positive 
    & 96.2\%\\
\midrule
    PSO
    & Part comedy, part drama, the movie winds up accomplishing neither in full, and leaves us feeling touched and amused by several moments and ideas, but nevertheless \textcolor{red}{dazzling} with the movie as a whole.
    & Positive 
    & 94.8\%\\
\midrule
    Pruthi
    & Part comedy, part drama, the movie winds up accomplishing neither in full, and leaves us feeling touched and amused by several moments and ideas, but neverthel\textcolor{red}{w}ss dissatisfied with the movie as a whole.  
    & Positive 
    & 95.5\%\\
\midrule
    TextFooler
    & Part comedy, part drama, the movie \textcolor{red}{gusty} up accomplishing neither in full, and leaves us feeling touched and amused by several moments and ideas, but nevertheless \textcolor{red}{incensed} with the movie as a whole. 
    & Positive 
    & 74.1\%\\
\midrule
    Genetic
    &  \textcolor{red}{Portions} comedy, part \textcolor{red}{theatrical}, the movie winds up \textcolor{red}{achieving nor} in full, and leaves us feeling touched and \textcolor{red}{tickled} by \textcolor{red}{diverse} moments and \textcolor{red}{thought}, but \textcolor{red}{however} dissatisfied with the movie as a whole.
    & Positive 
    & 58.3\%\\
\midrule
    TextBugger(TA)
    & Part comedy, part drama, the movie winds up accomplishing \textcolor{red}{nor} in full, and leaves us feeling touched and amused by several moments and ideas, but nevertheless \textcolor{red}{‘Å}issatisfied with the movie as a whole. 
    & Positive 
    & 53.3\%\\
\bottomrule
\end{tabular}
\caption{Attack Samples of SST against XLNet.}
\label{tab:attack_samples_sst}
\end{table*}

\begin{table*}[tb]
\small
\setlength\tabcolsep{3pt}
\center
\begin{tabular}{lm{9cm}cr}
\toprule
\textbf{Attack} & \textbf{Text} & \textbf{Label} & \textbf{Confidence}\\
\midrule
    Original
    & YOu moron, do not accuse me to go and study. The fools and used like condom people with small or with no brains are YOU! \dots
    & Toxic
    & 98.3 \%\\
\midrule
    BAE
    & YOu \textcolor{red}{suke}, do not accuse me to go and study. The fools and used like condom people with small or with no brains are YOU! \dots
    & Non-toxic
    & 78.5\%\\
\midrule
    DeepWordBug
    & \textcolor{red}{Yu mBron}, do not accuse me to go and study. The fools and used like condom people with small or with no brains are YOU! \dots
    & Non-toxic
    & 76.0\%\\
\midrule
    TextBugger(TA)
    & YOu \textcolor{red}{moon}, do not accuse me to go and study. The fools and used like condom people with small or with no brains are YOU! \dots
    & Non-toxic
    & 68.4\%\\
\midrule
    TextFooler
    & \textcolor{red}{tu cretin}, do not accuse me to go and study. The fools and used like condom people with small or with no brains are YOU! \dots
    & Non-toxic
    & 67.2\%\\
\midrule
    TextBugger(OA)
    &  Y\textcolor{red}{o}u moron , do not accuse me to go and study\textcolor{red}{ing}. The f\textcolor{red}{o}ls and \textcolor{red}{utilized loves} condom people with little or with no brains \textcolor{red}{z}re YOU!  \dots
    & Non-toxic
    & 91.1\%\\
\midrule
    Genetic
    & Y\textcolor{red}{o}u \textcolor{red}{cretin} , do not accuse me to go and study. The fools and \textcolor{red}{usage} like condom people with small or with no brains are \textcolor{red}{you}! \dots
    & Non-toxic
    & 54.4\%\\
\bottomrule
\end{tabular}
\caption{Attack Samples of Wikipedia against XLNet Example 1.}
\label{tab:attack_samples_wikipedia_1}
\end{table*}

\begin{table*}[tb]
\small
\setlength\tabcolsep{3pt}
\center
\begin{tabular}{lm{9cm}cr}
\toprule
\textbf{Attack} & \textbf{Text} & \textbf{Label} & \textbf{Confidence}\\
\midrule
    Original
    & Why do not the two of you, the mentally r***ded Jennygirl and the omnipotent Richie Rich, go together to Harrods and \dots when you come back after they would be kicking off your dipshit spinning incapable ass from the central perk. What a pity. Receive all my sympathy at one time. \dots
    & Toxic
    & 63.7 \%\\
\midrule
    Pruthi
    & Why do not the two of you, the mentally \textcolor{red}{regarded} Jennygirl and the omnipotent Richie Rich, go together to Harrods and \dots when you come back after they would be kicking off your dipshit spinning incapable ass from the central perk. What a pity. Receive all my sympathy at one time.\dots
    & Non-toxic
    & 82.1\%\\
\midrule
    PSO
    & Why do not the two of you, the mentally \textcolor{red}{checked} Jennygirl and the omnipotent Richie Rich, go together to Harrods and \dots when you come back after they would be kicking off your dipshit spinning incapable ass from the central perk. What a pity. Receive all my sympathy at one time. \dots
    & Non-toxic
    & 85.7\%\\
\midrule
    BAE
    & Why do not the two of you, the mentally r***ded Jennygirl and the omnipotent Richie Rich, go together to Harrods and \dots when you come back after they would be kicking off your \textcolor{red}{rella} spinning incapable ass from the central perk. What a pity. Receive all my sympathy at one time. \dots
    & Non-toxic
    & 92.2\%\\
\midrule
    TextFooler
    & Why do not the two of you, the mentally r***ded Jennygirl and the omnipotent Richie Rich, go together to Harrods and \dots when you come back after they would be kicking off your \textcolor{red}{cretin} spinning incapable ass from the central perk. What a pity. Receive all my sympathy at one time. \dots
    & Non-toxic
    & 91.2\%\\
\midrule
    Genetic
    &  Why do not the two of you, the mentally \textcolor{red}{backward} Jennygirl and the omnipotent Richie Rich, go together to Harrods and \dots when you come back after they would be kicking off your dipshit spinning incapable ass from the central perk. What a pity. Receive all my sympathy at one time. \dots
    & Non-toxic
    & 62.3\%\\
\midrule
    DeepWordBug
    &  Why do not the two of you, the mentally retarded Jennygirl and the omnipotent Richie Rich, go together to Harrods and \dots when you come back after they would be kicking off your dips\textcolor{red}{g}it spinning incapable ass from the central perk. What a pity. Receive all my sympathy at one time. \dots
    & Non-toxic
    & 93.7\%\\
\bottomrule
\end{tabular}
\caption{Attack Samples of Wikipedia against XLNet Example 2.}
\label{tab:attack_samples_wikipedia_2}
\end{table*}

\begin{table*}[tb]
\small
\setlength\tabcolsep{3pt}
\center
\begin{tabular}{m{6cm}m{1.6cm}m{6cm}m{1.6cm}}
\toprule
\textbf{Original Text} & \textbf{Original} & \textbf{Perturbed Text} & \textbf{Perturbed}\\
\midrule
    Watching Trouble Every Day , at least if you do n't know what 's coming , is like biting into what looks like a juicy , delicious plum on a hot summer day and coming away with your mouth full of rotten pulp and living worms .
    & Negative, 92.6\%
    & Watching Trouble Every Day , at least if you do n't know what 's coming , is like biting into what looks like a juicy , delicious plum on a hot summer day and coming away with your mouth full of \textcolor{red}{fresh} pulp and living worms .
    & Positive, 55.3\%\\
\midrule
    With a story inspired by the tumultuous surroundings of Los Angeles , where feelings of marginalization loom for every dreamer with a burst bubble , The Dogwalker has a few characters and ideas , but it never manages to put them on the same path .
    & Negative, 91.1\%
    & With a story inspired by the tumultuous surroundings of Los Angeles , where feelings of marginalization loom for every dreamer with a burst bubble , The Dogwalker has a few characters and ideas , but it \textcolor{red}{still} manages to put them on the same path .
    & Positive, 98.2\% \\
\midrule
    To imagine the life of Harry Potter as a martial arts adventure told by a lobotomized Woody Allen is to have some idea of the fate that lies in store for moviegoers lured to the mediocrity that is Kung Pow : Enter the Fist .
    & Negative, 94.0\%
    & To \textcolor{red}{experience} the life of Harry Potter as a martial arts adventure told by a \textcolor{red}{mad} Woody Allen is to have some idea of the fate that lies in store for moviegoers lured to the mediocrity that is Kung Pow : Enter the Fist .
    & Positive, 51.1\%\\
\midrule
    This is n't a narrative film -- I do n't know if it 's possible to make a narrative film about September 11th , though I 'm sure some will try -- but it 's as close as anyone has dared to come .
    & Positive, 63.8\%
    & This is n't a narrative film -- I \textcolor{red}{do t} know if it 's possible to make a narrative film about September 11th , though I 'm sure some will try -- but it 's as close as anyone has dare to come .
    & Negative, 66.6\%\\
\midrule
    Though Mama takes a bit too long to find its rhythm and a third-act plot development is somewhat melodramatic , its ribald humor and touching nostalgia are sure to please anyone in search of a Jules and Jim for the new millennium .
    & Positive, 98.2\%
    & \textcolor{red}{as} Mama takes a bit too long to find its rhythm and a third-act plot development is somewhat \textcolor{red}{questionable} , its ribald humor and touching nostalgia are sure to \textcolor{red}{deter} anyone in search of a Jules and Jim for the new millennium .
    & Negative, 54.7\%\\
\bottomrule
\end{tabular}
\caption{Attack Samples of SST Attacked by BAE against RoBERTa. The ``Original" column contains the original label and the prediction confidence. The ``Perturbed" column contains the perturbed label and the perturbation confidence.}
\label{tab:attack_samples_sst_multi}
\end{table*}

\begin{table}[tb]
\small
\setlength\tabcolsep{3pt}
\center
\begin{tabular}{l cccccccccccc}
\toprule
\multirow{2}{*}{\bf Feature}
  & \multicolumn{12}{c}{\textbf{Attack}} \\
\cmidrule(lr){2-13}
 & Clean & BAE & DWB & FG & Gn.* & HF* & IGA & Pr. & PSO & TB & TF & VIP* \\
\midrule
TP AVG WORD LENGTH
     & 1.1 & 1.0 & 2.1
     & 2.4 & 2.9 & 3.6
     & 4.2 & 4.7 & 5.4
     & 5.8 & 6.4 & 6.5
\\
TP BERT
     & 6.3 & 3.7 & 9.6
     & 12.8 & 15.6 & 18.0
     & 20.9 & 24.6 & 28.1
     & 30.8 & 33.9 & 34.6
\\
TP FIRST WORD LOWERCASE
     & 0.2 & 0.2 & 0.3
     & 0.3 & 0.5 & 0.7
     & 0.7 & 0.7 & 0.7
     & 0.7 & 0.7 & 0.7
\\
TP NUM ALPHA CHARS
     & 0.0 & 0.0 & 0.0
     & 0.2 & 0.2 & 0.2
     & 0.2 & 0.2 & 0.7
     & 0.7 & 1.3 & 1.4
\\
TP NUM CASED LETTERS
     & 0.5 & 0.2 & 0.9
     & 1.6 & 1.7 & 4.2
     & 4.5 & 4.6 & 4.9
     & 5.9 & 7.0 & 7.0
\\
TP NUM CASED WORD SWITCHES
     & 0.0 & 0.0 & 0.0
     & 0.2 & 0.2 & 1.3
     & 1.7 & 1.7 & 1.7
     & 1.9 & 2.1 & 2.1
\\
TP NUM CHARS
     & 0.0 & 0.0 & 0.0
     & 0.0 & 0.0 & 0.4
     & 0.4 & 0.4 & 0.4
     & 0.4 & 0.4 & 0.5
\\
TP NUM DIGITS
     & 0.0 & 0.0 & 0.5
     & 0.5 & 0.5 & 0.5
     & 0.5 & 0.8 & 0.8
     & 0.8 & 1.0 & 1.0
\\
TP NUM LOWERCASE AFTER PUNCTUATION
     & 0.2 & 0.0 & 0.2
     & 0.2 & 0.3 & 0.3
     & 0.3 & 0.4 & 0.4
     & 0.5 & 0.5 & 0.5
\\
TP NUM SINGLE LOWERCASE LETTERS
     & 0.6 & 0.4 & 1.2
     & 1.2 & 1.2 & 1.2
     & 1.4 & 1.6 & 1.6
     & 1.8 & 1.9 & 1.9

\\
TP NUM MIXED CASE WORDS
     & 0.4 & 0.1 & 2.6
     & 2.9 & 3.2 & 4.7
     & 5.0 & 5.9 & 6.0
     & 7.3 & 8.2 & 8.2
\\
TP NUM MULTI SPACES
     & 0.0 & 0.0 & 0.1 
     & 0.1 & 0.1 & 0.1
     & 0.1 & 0.2 & 0.2
     & 0.2 & 0.2 & 0.2
\\
TP NUM NON ASCII
     & 0.2 & 0.0 & 4.2
     & 4.2 & 4.2 & 4.2
     & 4.5 & 5.2 & 5.2
     & 5.2 & 6.7 & 8.9
\\
TP NUM PUNCTUATION
     & 0.0 & 0.0 & 0.0
     & 0.0 & 0.0 & 0.0
     & 0.0 & 0.0 & 0.0
     & 0.0 & 0.0 & 0.0
\\
TP NUM WORDS
     & 1.1 & 1.1 & 2.7
     & 3.0 & 3.0 & 4.7
     & 5.3 & 6.8 & 6.8
     & 6.8 & 10.0 & 10.0
\\
LM PERPLEXITY
     & 0.2 & 0.2 & 0.2
     & 0.3 & 0.3 & 0.3
     & 0.4 & 0.5 & 0.5
     & 0.6 & 0.8 & 0.8
\\
LM PROBA AND RANK
     & 1.5 & 1.1 & 2.2
     & 3.2 & 3.9 & 4.3
     & 5.0 & 5.8 & 6.2
     & 6.5 & 7.0 & 7.0
\\
TM ACTIVATION
     & 9.7 & 6.2 & 15.4
     & 19.5 & 23.5 & 28.6
     & 33.4 & 39.4 & 44.1
     & 49.8 & 54.6 & 57.3
\\
TM GRADIENT
     & 9.3 & 5.7 & 13.1
     & 16.9 & 20.9 & 23.0
     & 27.7 & 33.0 & 36.0
     & 39.9 & 44.9 & 45.1
\\
TM POSTERIOR
     & 0.0 & 0.0 & 0.0
     & 0.0 & 0.0 & 0.0
     & 0.0 & 0.0 & 0.0
     & 0.0 & 0.0 & 0.0
\\
TM SALIENCY
     & 0.2 & 0.2 & 0.2
     & 0.3 & 0.4 & 0.8
     & 1.2 & 2.1 & 2.3
     & 2.9 & 3.0 & 3.0
\\
\bottomrule
\end{tabular}
\caption{All Features (and their contributions, see Eq.~\ref{eu_eqn}) for each attack method for the L-TLC~(RoBERTa) labeling model on the Hatebase dataset.}
\label{tab:all_features}
\end{table}

\end{document}